\documentclass{ecai} 
\usepackage{latexsym}
\usepackage{amssymb}
\usepackage{amsmath}
\usepackage{amsthm}
\usepackage{booktabs}
\usepackage{enumitem}
\usepackage{graphicx}
\usepackage{color}
\usepackage{times}
\usepackage{soul}
\usepackage{url}
\usepackage[hidelinks]{hyperref}
\usepackage[utf8]{inputenc}
\usepackage[small]{caption}
\usepackage{graphicx}
\usepackage{amsmath}
\usepackage{amsfonts}
\usepackage{amsthm}
\usepackage{booktabs}
\usepackage{algorithm}
\usepackage{algorithmic}
\usepackage{multirow}
\usepackage{subcaption}
\usepackage[switch]{lineno}
\usepackage{ amssymb }

\newcommand{\BibTeX}{B\kern-.05em{\sc i\kern-.025em b}\kern-.08em\TeX}

\begin{document}

%%%%%%%%%%%%%%%%%%%%%%%%%%%%%%%%%%%%%%%%%%%%%%%%%%%%%%%%%%%%%%%%%%%%%%%%

\begin{frontmatter}

%%% Use this command to specify your submission number.
%%% In doubleblind mode, it will be printed on the first page.

\paperid{1535} 

%%% Use this command to specify the title of your paper.

\title{Learning Backdoors for Mixed Integer Linear Programs with Contrastive Learning}

%%% Use this combinations of commands to specify all authors of your 
%%% paper. Use \fnms{} and \snm{} to indicate everyone's first names 
%%% and surname. This will help the publisher with indexing the 
%%% proceedings. Please use a reasonable approximation in case your 
%%% name does not neatly split into "first names" and "surname".
%%% Specifying your ORCID digital identifier is optional. 
%%% Use the \thanks{} command to indicate one or more corresponding 
%%% authors and their email address(es). If so desired, you can specify
%%% author contributions using the \footnote{} command.

% \author[]{Paper ID: 1535}

\author[A]{\fnms{Junyang}~\snm{Cai}\thanks{Correspondence to: Junyang Cai <caijunya@usc.edu>}}
\author[B]{\fnms{Taoan}~\snm{Huang}}
\author[C]{\fnms{Bistra}~\snm{Dilkina}}
% \author[B]{\fnms{Second}~\snm{Author}\orcid{....-....-....-....}\footnotemark}
% \author[B,C]{\fnms{Third}~\snm{Author}\orcid{....-....-....-....}} 

\address[A,B,C]{University of Southern California}
% \address[B]{Short Affiliation of Second Author and Third Author}
% \address[C]{Short Alternate Affiliation of Third Author}

%%% Use this environment to include an abstract of your paper.

\begin{abstract}
% Motivation
Many real-world problems can be efficiently modeled as Mixed Integer Linear Programs (MILPs) and solved with the Branch-and-Bound method. Prior work has shown the existence of MILP backdoors, small sets of variables such that prioritizing branching on them when possible leads to faster running times.
% Problem
However, finding high-quality backdoors that improve running times remains an open question.
% Previous work
%Recent developments include learning-based methods for identifying effective heuristics in combinatorial optimization problems.
Previous work learns to estimate the relative solver speed of randomly sampled backdoors through ranking and then decide whether to use the highest-ranked backdoor candidate. 
% Approach
In this paper, we utilize the Monte-Carlo tree search method to collect backdoors for training, rather than relying on random sampling, and adapt a contrastive learning framework to train a Graph Attention Network model to predict backdoors.
% Conclusion
Our method, evaluated on several common MILP problem domains, demonstrates performance improvements over both Gurobi and previous models.
\end{abstract}

\end{frontmatter}

%%%%%%%%%%%%%%%%%%%%%%%%%%%%%%%%%%%%%%%%%%%%%%%%%%%%%%%%%%%%%%%%%%%%%%%%

\section{Introduction}
Many real-world problems, such as path planning~\cite{pohl1970heuristic}, scheduling~\cite{floudas2005mixed}, and network design \cite{dilkina2010solving,huang2020enhancing} problems, are combinatorial optimization (CO) problems and generally NP-hard to solve. Many CO problems can be formulated as Mixed Integer Linear Programs (MILPs), which aim to optimize a linear objective function over both continuous and integer variables in the feasible region defined by linear constraints. MILPs can be solved via Branch-and-Bound~\cite{land2010automatic},  which repeatedly solves a linear program (LP) relaxation of the problem and selects one of the integer variables that is fractional in the LP relaxation solution to branch on and creates new subproblems until all the integrality constraints are satisfied. Commercial MILP solvers, such as Gurobi~\cite{gurobi}, heavily rely on Branch-and-Bound algorithms and are enhanced by various powerful heuristics.

Backdoors, initially introduced for Constraint Satisfaction Problems ~\cite{williams2003backdoors}, are generalized to MILPs~\cite{dilkina2009backdoors}, where MILP "strong backdoors" are defined as subsets of integer variables such that branching on only them yields an optimal integral solution. ~\cite{fischetti2011backdoor} report speedup in MILP solving times by prioritizing backdoor variables in branching. Instead of node selection, variable selection happens during the branching; we assign branching priority for backdoor variables before the start of the tree search. Previously,~\cite{dilkina2009backdoors} proposed a sampling method for finding backdoors by randomly selecting from a subset of variables based on the fractionality of their LP relaxation solutions. More recently,~\cite{khalil2022finding} developed a Monte-Carlo tree search (MCTS) approach for finding backdoors. However, the performance of backdoors in a problem domain or even a specific instance varies and it is hard to identify backdoors that improve solving time efficiently.

Because of the advantage of the machine learning (ML) model to learn from complex historical distribution, learning-based approaches have been successfully used in combinatorial optimization, specifically speeding up solving MILPs. There has been an increased interest in data-driven heuristic designs for MILP for various decision-making in Branch-and Bound \cite{zhang2023survey} such as node selection and branch variables selection. Those works treat MILP solvers as a white box; they repeatedly make decisions at each node or every branching decision during solving time, which requires special implementations of open-source solvers, e.g., SCIP. While our backdoor prediction treats the solvers as a black box, we only make one branching priority decision ahead of the solving, which gives us the advantage of being easily applied to all solvers. ~\cite{ferber2021learning} is the first to use the learning-based approach to find backdoors and propose a two-step, learning-based model. In the first step, a scorer model is trained with a pairwise ranking loss to rank candidate backdoors from random sampling by solver runtime. In the second step, a classifier model is trained with cross-entropy loss to determine whether to use the best-scoring backdoors or the default MILP solver. Recently, contrastive learning has emerged as an effective approach for other tasks in CO, such as solving satisfiability problems (SAT)  ~\cite{duan2022augment} and improving large neighborhood search for MILPs~\cite{huang2023searching}, based on learning representations by contrasting positive samples (e.g., high-quality solutions or heuristic decisions) against negative ones. 

In this paper, we introduce a novel learning framework for predicting effective backdoors. Instead of using sampling methods \cite{ferber2021learning}, we employ the Monte-Carlo tree search (MCTS)~\cite{khalil2022finding} for data collection that allows us to find higher-quality backdoors compared to previous methods \cite{ferber2021learning}. High-quality backdoors found by the MCTS method are used as positive samples for contrastive learning. We also collect backdoor samples that cause performance drops as negative samples. We represent MILPs as bipartite graphs with variables and constraints being the two sets of nodes in the graphs ~\cite{gasse2019exact} and parameterize the ML model using a graph attention network~\cite{velivckovic2017graph}. We use a contrastive loss that encourages the model to predict backdoors similar to the positive samples but dissimilar to the negative ones ~\cite{oord2018representation} and use a greedy selection process to predict the most beneficial backdoors at test time. Compared to the previous model, contrastive learning is more efficient in training and more deterministic in evaluation. We conduct empirical experiments on several common problem domains, including Generalized Independent Set Problem (GISP)~\cite{hochbaum1997forest}, Set Cover Problem (SC)~\cite{balas1980set}, Combinatorial Auction (CA)~\cite{leyton2000towards}, Maximal Independent Set (MIS)~\cite{tarjan1977finding}, Facility Location(FC)~\cite{cornuejols1991comparison}, and Neural Network Verification(NN) ~\cite{nair2020solving}. Our method achieves faster solve times than Gurobi and the state-of-the-art scorer+classifier model proposed in~\cite{ferber2021learning}.

\section{Background}
In this section, we first define MILPs and then introduce backdoor for MILP solving.
\subsection{Mixed Integer Linear Programming}
A Mixed Integer Linear Programming (MILP) is defined as 
\begin{align}
    \min\{c^Tx\mid Ax\leq b, x\in\mathbb{R}^{n}, x_j \in \{0,1\} \forall j \in I\} \nonumber
\end{align}
where $A \in \mathbb{R}^{m\times n}, b \in \mathbb{R}^m, c \in \mathbb{R}^n$, and the non-empty set $I \subseteq {1, ..., n}$ indexes the binary variables.  A solution to the MILP is a feasible assignment of values to the variables. In this paper, we focus on the formulation above that consists of continuous and binary variables due to MCTS data collection limited to this setting, but our contrastive learning framework can still be applied to MILP with non-binary integer variables.

\subsection{Backdoors for MILP}
Given a MILP instance defined by the tuple $(A,b,c,I)$, a ``strong backdoor" of size $K \ll |I|$ is a set of integer variables $B$, $|B| = K$, $B \subset I$ such that branching exclusively on variables from $B$ results in a provably optimal solution or a proof of infeasibility. 
% While the term "backdoor" is defined as a set of integer variables such that prioritize branching on them leads to faster solving time. Throughout the paper, we use "candidate backdoors" to describe a set of integer variables currently under evaluation, which may or may not yield improvements in solving time.
The order in which decision variables are considered for branching can affect the performance of the solver's primal heuristics, which in turn affects pruning in Branch-and-Bound. We refer to \cite{dilkina2009backdoors, fischetti2011backdoor} for a detailed discussion of why order matters in practice in MILP solvers. To distinguish from ``strong backdoors'', we introduce term ``backdoor'' to denote a set of integer variables where prioritizing branching on them reduces solving time. Throughout the paper, we use ``candidate backdoors'' to describe a set of integer variables currently under evaluation, which may or may not yield improvements in solving time.

\section{Related Work}
In this section, we summarize related work on backdoors for CO problems, learning to solve MILPs, and contrastive learning for CO problems.

\subsection{Backdoors For Combinatorial Problems}
Backdoors are first introduced by~\cite{williams2003backdoors} for SAT solving and, over the years, many approaches are proposed to find backdoors to improve SAT solving ~\cite{paris2006computing,kottler2008computation}. Observing the connection between SAT and MILP,~\cite{dilkina2009backdoors} generalize the concept of backdoors and propose two forms of random sampling to find backdoors: uniform sampling and biased sampling to favor ones that are more fractional in the LP relaxation solutions, while~\cite{fischetti2011backdoor} propose another strategy that formulates finding backdoors as a Set Covering Problem. Recently,~\cite{khalil2022finding} contributed to a third strategy for finding backdoors using Monte-Carlo tree search, which balances exploration and exploitation by design. Utilizing a reward function based on the tree weight in the Branch-and-Bound tree, MCTS approximates the backdoor's strength without the need to solve time-consuming MILP instances. However, the quality of the backdoors it finds varies and it is also time-consuming (5+ hours) for MCTS to find backdoors, which opens the floor for learning-based techniques to find backdoors ~\cite{ferber2021learning}.

\subsection{Machine Learning for MILPs}
Several studies have applied ML to improve heuristic decisions in Branch-and-Bound, such as which node to select next to expand \cite{he2014learning,song2018learning,labassi2022learning}, which variable to branch on next \cite{khalil2016learning,gasse2019exact,gupta2020hybrid}, which cut to select next \cite{tang2020reinforcement,paulus2022learning}, which primal heuristics to run next \cite{khalil2017learning,chmiela2021learning}. Another line of work focuses on improving speed to find primal solutions. \cite{song2020general,sonnerat2021learning,huang2023searching} use ML to iteratively decide which subsets of variables to re-optimize for a given solution to a MILP. \cite{han2022gnn,nair2020solving} learn to predict high-quality solutions to MILPs with supervised learning. 

\subsection{Contrastive Learning for CO}
Contrastive Learning is a discriminative approach that aims at grouping similar samples closer and dissimilar samples far from each other~\cite{jaiswal2020survey}. It has been successfully applied in both computer vision area~\cite{hjelm2018learning} and natural language processing area~\cite{logeswaran2018efficient}. The work of contrastive learning on graph representation has also been extensively studied~\cite{you2020graph}, but it has not been explored much for CO problem domains. ~\cite{mulamba2020contrastive} derive a contrastive loss for decision-focused learning to solve CO problems with uncertain inputs. ~\cite{duan2022augment} use contrastive pre-training to learn good representations for the boolean satisfiability problem. ~\cite{huang2023searching} use contrastive learning to learn efficient and effective destroy heuristics in large neighborhood search for MILPs. \cite{huang2024contrastive} use contrastive learning to predict optimal solutions to MILPs in the predict-and-search algorithm. 

\begin{figure*}
    \centering
    \includegraphics[width=1.02\linewidth]{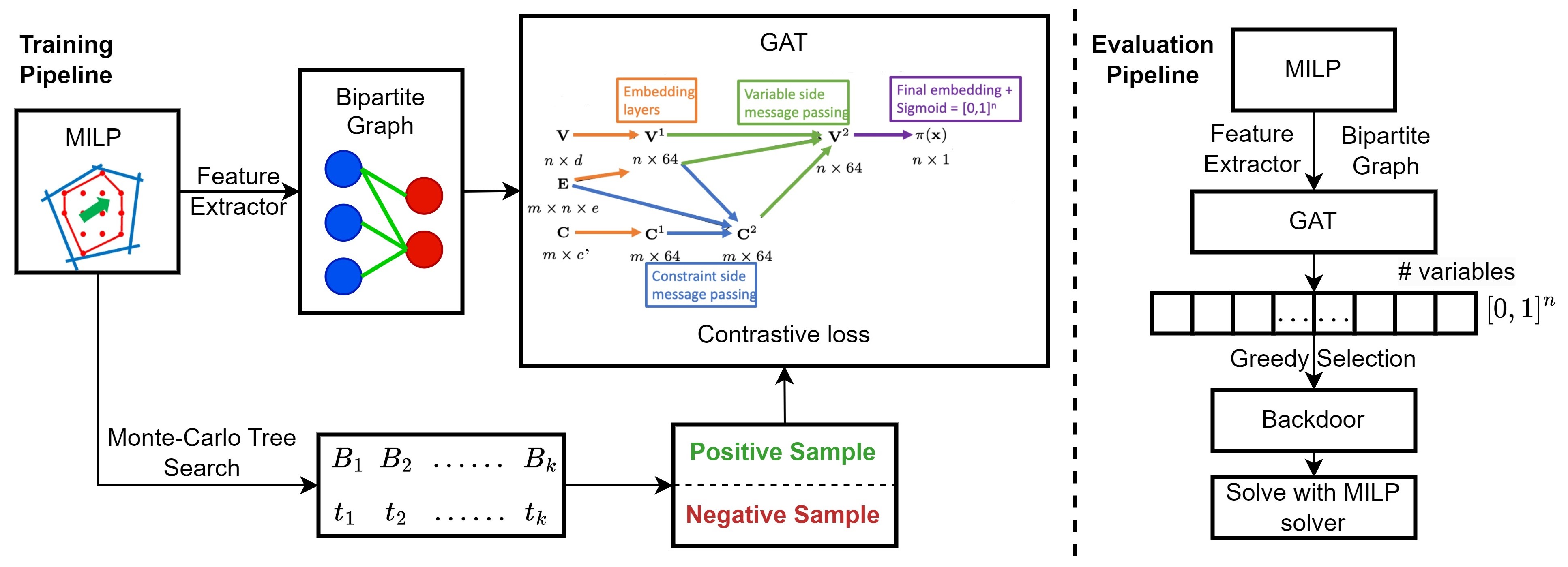}
    \caption{This figure illustrates the comprehensive training and evaluation phases for learning backdoors for MILPs with contrastive learning. In the training phase, we employ MCTS to gather backdoor samples on training instances and collect positive and negative samples. Simultaneously, a feature extractor transforms the MILP instances into a bipartite graph and collects features of the MILPs. The policy is represented by a graph attention network (GAT) and trained with a contrastive loss. In the evaluation phase, the MILP instance is converted into a bipartite graph as input for the GAT, generating a score vector as output. We greedily select variables with the highest score to obtain backdoors, which are solved using a MILP solver.\\}
    \label{fig:pipeline}
\end{figure*}

\section{Contrastive Learning for Backdoor Prediction} 
Our goal is to learn a policy that takes as input a MILP instance and outputs a backdoor to speed up solving time. In this section, we introduce how we prepare data for contrastive learning, the policy network, and the contrastive loss used in training. Finally, we introduce how the learned policy is used in predicting backdoors. Figure \ref{fig:pipeline} shows the full training and evaluation pipeline. 

\subsection{Data Collection} \label{3.1}
Diverging from previous approaches for finding or predicting backdoors \cite{ferber2021learning,fischetti2011backdoor,dilkina2009backdoors}, we incorporate contrastive learning to predict backdoors. The key challenge lies in determining high-quality positive samples and similar negative samples for effective training. a MILP instance can be represented as a tuple $P = (A,b,c,I)$, where $A,b,c$ are the coefficients of the constraints or objectives and $I$ is the set of integer variables. Our objective is to collect a training dataset $D$. Specifically, for every training instance $P$, we collect positive samples $S_p$ and negative samples $S_n$ specific to $P$. 

% Utilizing a reward function based on the tree weight in the Branch-and-Bound tree, MCTS provides a reliable approximation of the strength of backdoor without the need to solve time-consuming MILP instances. Through empirical evaluation of the backdoor collected by MCTS, we observe that the tree weight function is not always proportional to the solving time. For model training purposes, more accurate data is required. 

We employ the Monte-Carlo Tree Search Algorithm proposed in~\cite{khalil2022finding} to collect candidate backdoors. MCTS will return a set of candidate backdoors with their associate tree weights in the Branch-and-Bound tree, which are approximations of the strength of backdoors. We select the top $k$ candidate backdoors $B_1, B_2, \dots, B_k$ with the highest tree weights, and solve the MILP $P$ with these candidate backdoors to obtain their runtimes $t_1, t_2, \dots, t_k$. When we solve a MILP with a backdoor, we set the branching priority for variables in the backdoor higher than other variables, i.e., we always branch on variables from the backdoor whenever there is one that needs to be branched on. Positive samples $S_p$ are chosen as $p$ shortest-runtime backdoors better than Gurobi and negative samples $S_n$ are chosen as $q$ longest-runtime backdoors worse than Gurobi. The approach ensures that the positive samples are backdoors with high MILP runtime performance. In contrast, the negative samples are similar to positive samples but have worse runtime performance, enabling the ML model to learn to discriminate between backdoors and non-backdoors effectively. In experiments, we set $k=50$ and $p=q=5$. Details of data collection design and sensitive analysis are provided in Appendix~\cite{CAI_Learning_Backdoors_for_2024}.

\subsection{Data representation and Policy network} \label{section:data}
We represent the MILP $P = (A,b,c,I)$ as a bipartite graph as in ~\cite{gasse2019exact}. The featured bipartite graph $P' = (G,V,C,E)$, with graph $G$ containing variable nodes and constraint nodes. Additionally, there is an edge $(i, j)$ in edge matrix $\epsilon$ between a variable node $i$ and a constraint node $j$ if variable $i$ appears in constraint $j$ with nonzero coefficient, i.e., $A_{ji} \neq 0$. Constraint, variable, and edge features are represented as matrices $V \in \mathbb{R}^{n\times d}, C \in \mathbb{R}^{m\times c'}, E \in \mathbb{R}^{m \times n \times e}$. The bipartite graph representation ensures the MILP encoding is invariant to variable and constraint permutation. Additionally, we can use a variety of predictive models designed for variable-sized graphs, enabling deployment on problems with varying numbers of variables and constraints. We use features proposed in~\cite{gasse2019exact}, which include 15 variable features (e.g., variable types, coefficients, upper and lower bound, root LP related features), 4 constraint features (e.g., constant, sense), and 1 edge feature for coefficients. 

We learn a policy $\pi$ represented by a Graph Attention Network (GAT)~\cite{brody2021attentive}, which takes the bipartite graph $P'$ as input and outputs a score vector with one score per decision variable. To increase the modeling capacity and to manipulate node interactions proposed by our architecture, we use embedding layers in the network to first adjust the size of feature embeddings to $V^1 \in \mathbb{R}^{n\times L}, C^1 \in \mathbb{R}^{m\times L}, E^1 \in \mathbb{R}^{|\epsilon|\times L}$ accordingly. Then, GAT performs two rounds of message passing. In the first round, each constraint node in $C^1$ attends to its neighbors using an attention structure with $H$ attention heads to get update constraint embeddings $C^2$. Similarly, each variable node in $V^1$ attends to its neighbors to get updated variable embeddings $V^2$ with another set of attention weights in the second round. Finally, the network applies a multi-layer perceptron with a sigmoid function to obtain a score between 0 and 1 for each variable. In experiments, layer size $L$ and number of attention heads $H$ are set to 64 and 8. Full details of the network architecture are provided in Appendix~\cite{CAI_Learning_Backdoors_for_2024}. 

\begin{table*}[ht!]
\centering
\caption{Average numbers of variables, constraints, and average Gurobi solving time of 100 test instances from each problem domain. For MILP domains, \#variables is in the format of \# binary variables + \# continuous variables.\\}
\begin{tabular}{l|rrrr|rrrr|rr}
\toprule
              & \multicolumn{4}{c}{Small ILP Domains} & \multicolumn{4}{|c}{Large ILP Domains} & \multicolumn{2}{|c}{MILP Domains} \\ 
\midrule
Name   & GISP-S   & SC-S   & CA-S   & MIS-S   & GISP-L   & SC-L   & CA-L   & MIS-L  & FC & NN   \\ 
\midrule
\#Variables  &988      & 1,000   & 750    & 1,250    & 1,316     & 1,000   & 1,000   & 1,500   & 100 + 20,000  & 167 + 6968 \\ 
\#Constraints  &3,253     & 1,200   & 282    & 3,946    & 4,571     & 1,500   & 377    & 5,941  & 40,000 & 6520  \\
Gurobi runtime (s)   & 633      & 171    & 177    & 218     & 3,363     & 1,195   & 1,213   & 759   & 46.7 & 9.24  \\
\bottomrule
\end{tabular}
\label{table:a}
\end{table*}

\subsection{Training with Contrastive Loss and Applying Learned Policy} \label{3.3}
The model is trained with a contrastive loss to score each decision variable by learning to emulate superior backdoors and avoid inferior ones. Given a set of MILP instances for training, let $D = (P', S_p, S_n)$ be the set of one training data where $P'$ is the bipartite representation of MILP instance $P$, $S_p$ is the set of positive samples and $S_n$ is the set of negative samples associated with $P$. With similarity measured by dot products, we use the InfoNCE \cite{oord2018representation} contrastive loss  $L(P', S_p, S_n) = $
\begin{equation}
    {\sum_{(P',S_p,S_n)\in D}\frac{-1}{|S_p|}\sum_{a\in S_p}\log \frac{\exp(a^T\pi(P')/\tau)}{\sum_{a'\in S_n\cup\{a\}\exp(a'^T\pi(P')/\tau)}}}\nonumber
\end{equation} 
to train the policy $\pi$, where $\tau$ is a temperature hyperparameter set to $0.07$ in experiment following ~\cite{huang2023searching}. By choosing faster solving time backdoors as positive samples and those similar to the positive ones but with lower quality as negative samples, the model learns to generate embeddings closer to positive samples and further away from negative ones.

\begin{figure} [ht]
 \begin{subfigure}{0.24\textwidth}
     \includegraphics[width=\textwidth]{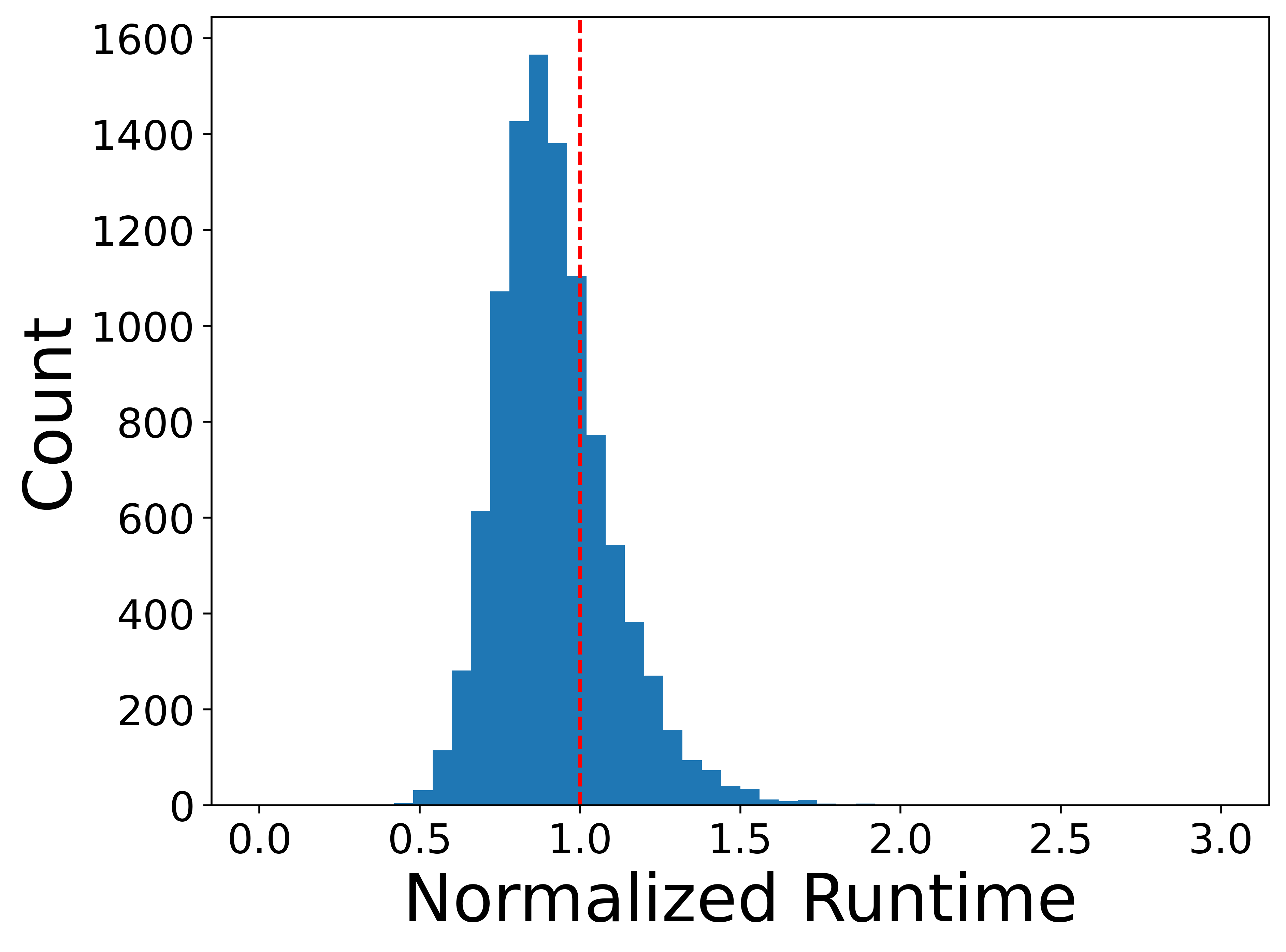}
     \caption{GISP-S}
     \label{fig:a}
 \end{subfigure}
 \begin{subfigure}{0.24\textwidth}
     \includegraphics[width=\textwidth]{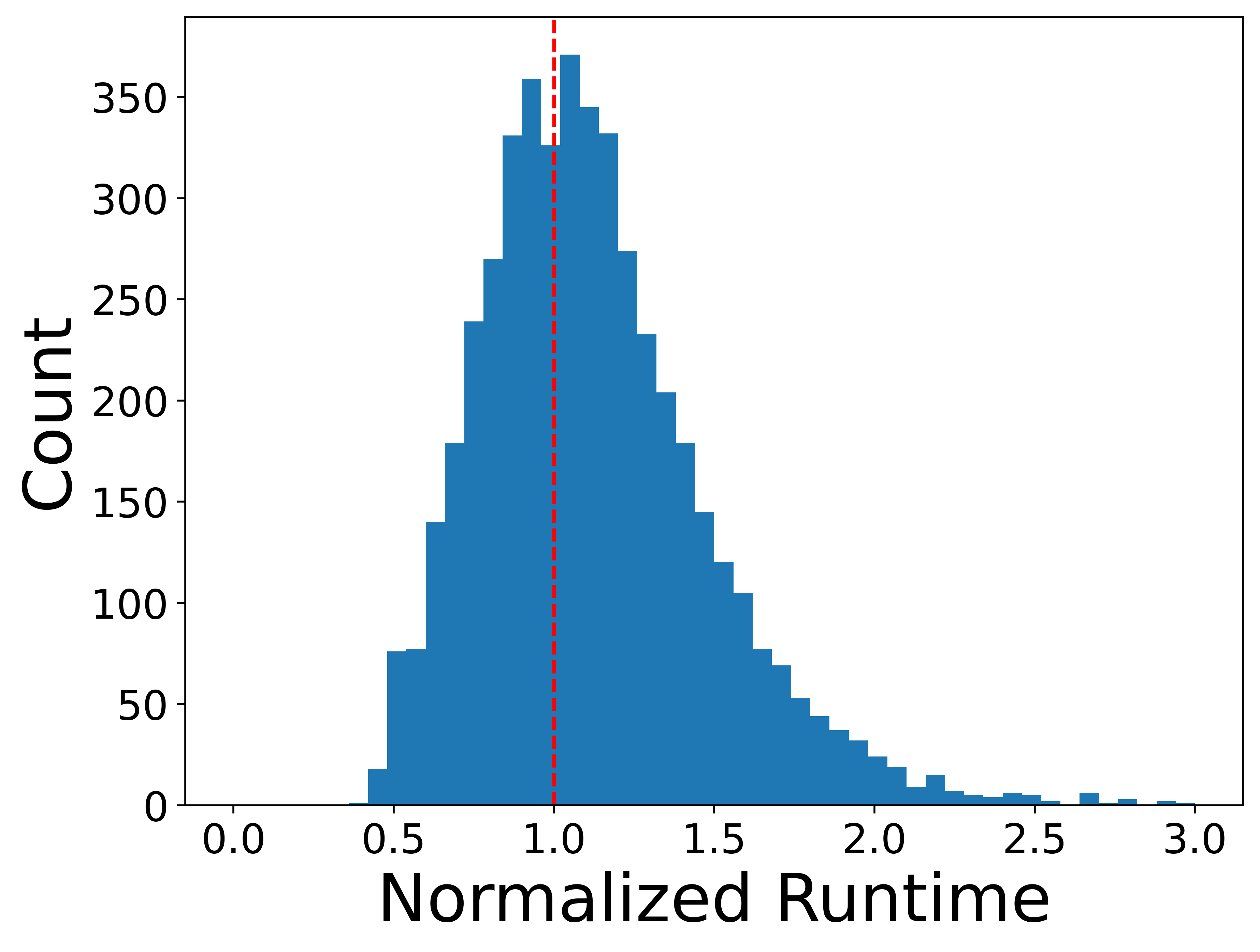}
     \caption{SC-S}
     \label{fig:b}
 \end{subfigure}
\\
\\
 \begin{subfigure}{0.24\textwidth}
     \includegraphics[width=\textwidth]{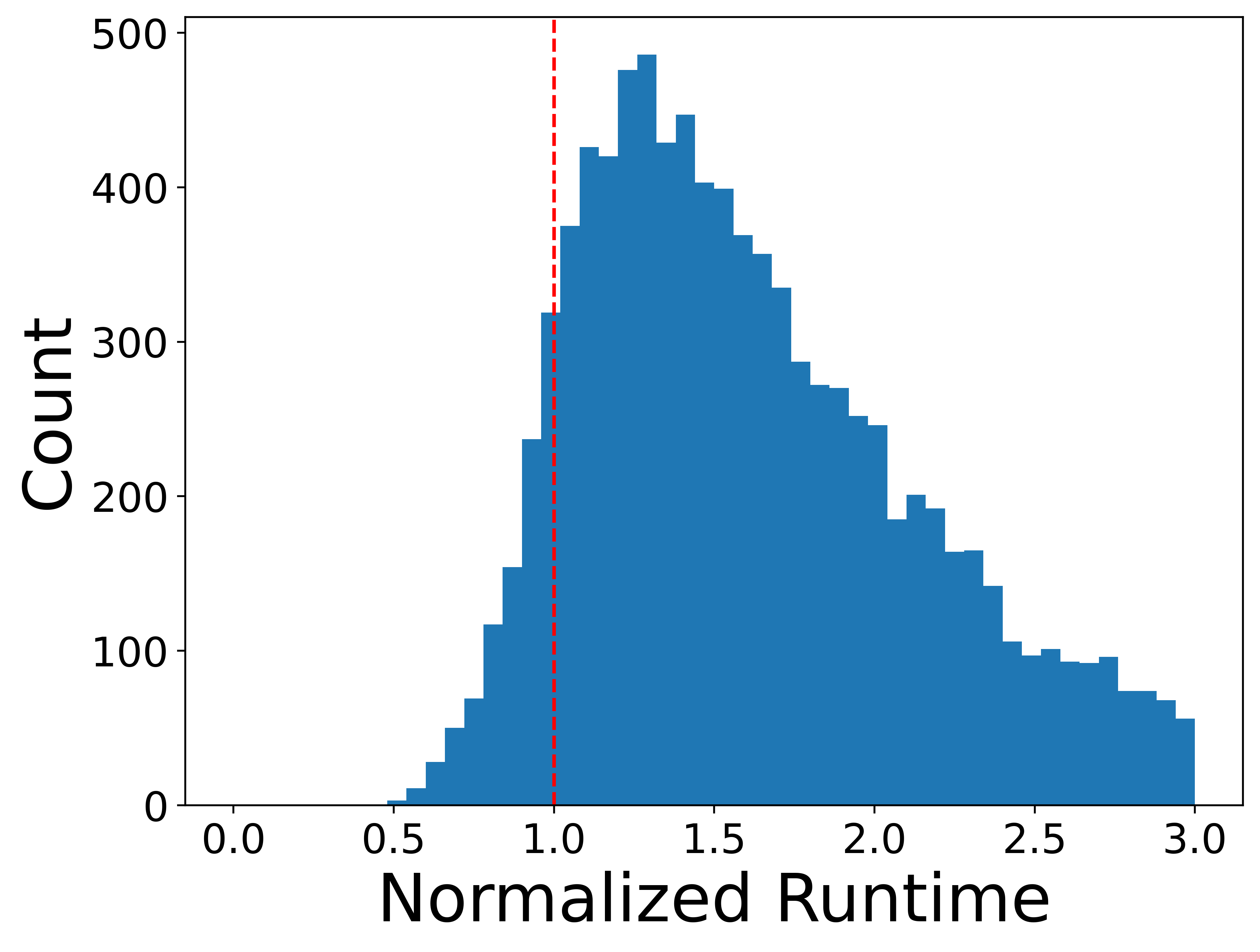}
     \caption{CA-S}
     \label{fig:c}
 \end{subfigure}
 \begin{subfigure}{0.24\textwidth}
     \includegraphics[width=\textwidth]{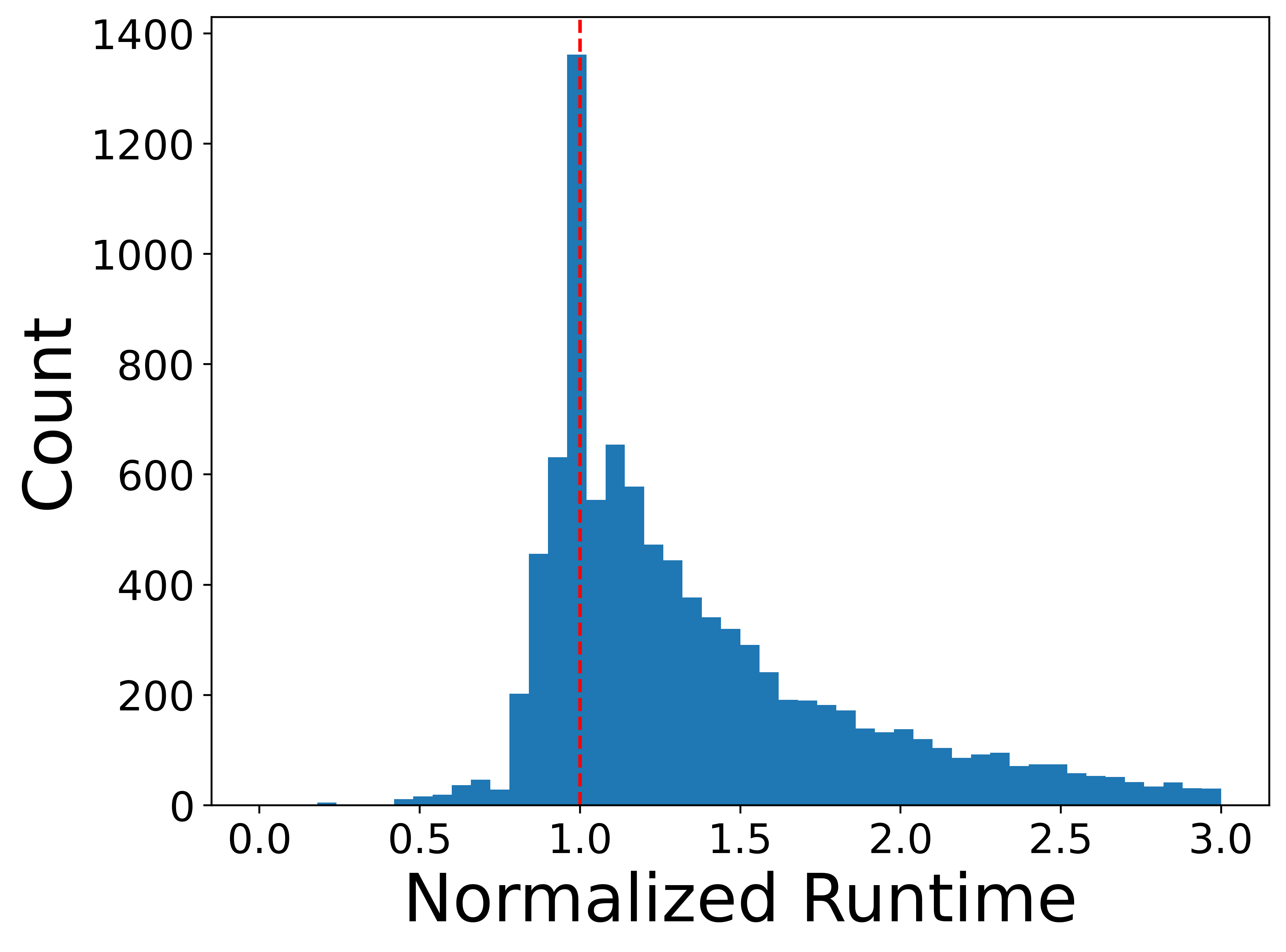}
     \caption{IS-S}
     \label{fig:c}
 \end{subfigure}
 \\
 \\
 \begin{subfigure}{0.24\textwidth}
     \includegraphics[width=\textwidth]{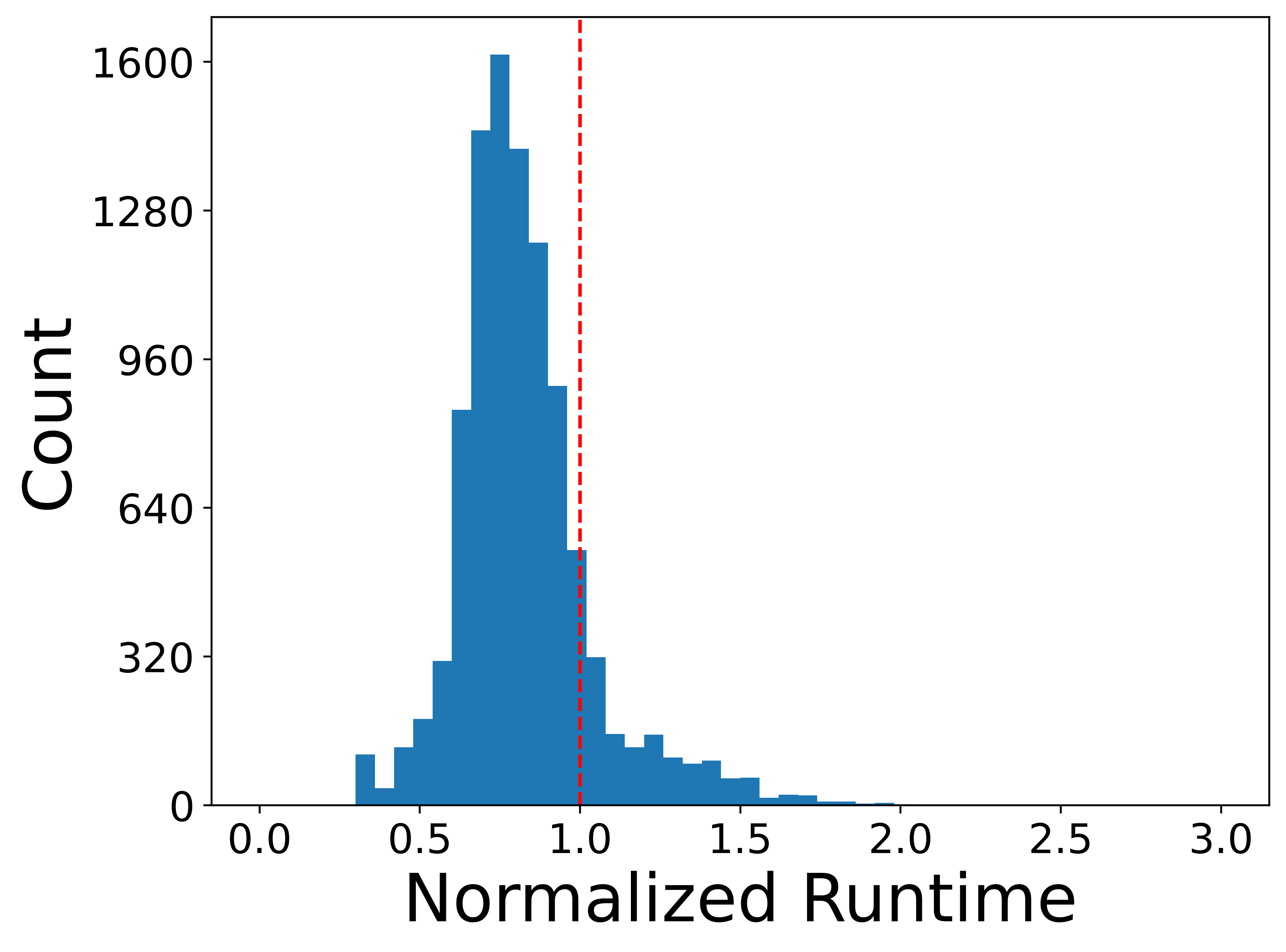}
     \caption{FC}
     \label{fig:d}
 \end{subfigure}
 \hfill
 \begin{subfigure}{0.24\textwidth}
     \includegraphics[width=\textwidth]{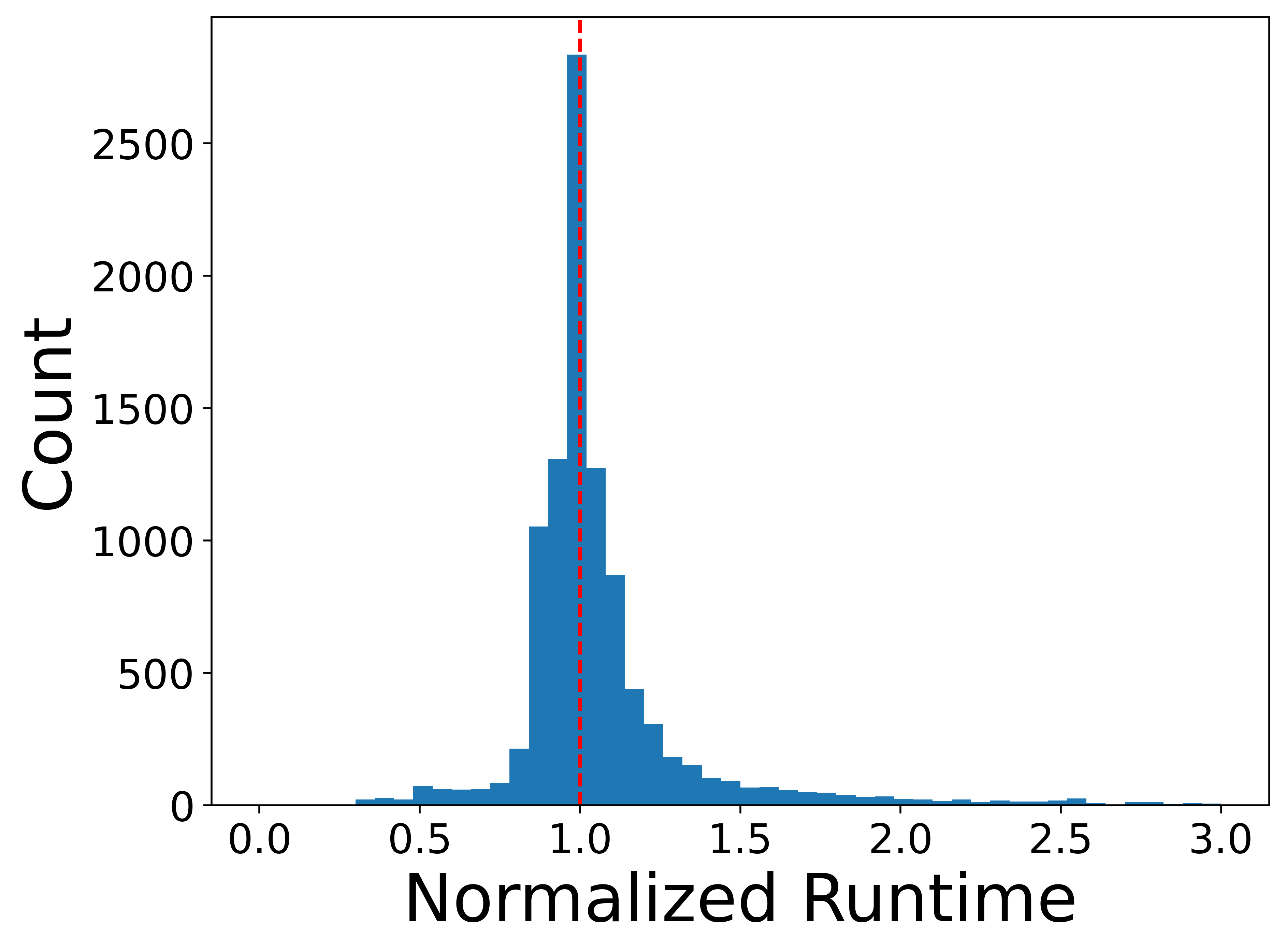}
     \caption{NN}
     \label{fig:c}
 \end{subfigure}
 \\
 \caption{Histograms illustrating the normalized runtime distributions of candidate backdoors for six distinct problem domains: (a) GISP-S, (b) SC-S, (c) CA-S, (d) IS-S, (e) FC, and (f) NN. The normalized runtime is calculated as the ratio of the solve time with the candidate backdoor to the original solve time without it. The red vertical line at 1.0 marks the threshold where the candidate backdoor's performance equals the original solve time. Values to the left of this line indicate instances where the candidate backdoor resulted in a faster solve time, while values to the right indicate a slower solve time than the original.\\\\}
  \label{fig:bd}
\end{figure}

During testing, given a MILP instance, we use the same data representation introduced in section \ref{section:data} to convert it to a bipartite graph and use the graph as the input to GAT. Then, GAT outputs a score vector with one score for each variable. We greedily select the binary variables with the highest scores as the predicted backdoors by user-defined backdoor size. For the scorer model in~\cite{ferber2021learning}, predicting the backdoor differs from our contrastive learning model. They first sample 50 candidate backdoors with the predefined backdoor size using the random biased sampling method \cite{dilkina2009backdoors} and input each backdoor to the scorer model to get an output score. They select the candidate backdoor with the highest score as the predicted backdoor. The backdoor predicted by the scorer model contains strong randomness within the 50 candidate backdoors it samples, thus causing the predicted one to have highly variate performance results. Our contrastive learning model is deterministic based on greedy selection and can also output a backdoor with any size requested by the users. Finally, we assign the branching priority for variables in the backdoor before the start of the tree search in the MILP solver and then collect statistics of solving the problem to optimality for later evaluation.

\section{Empirical Evaluation}
In this section, we introduce problem domains and instance sizes for evaluation. We explain the experiment setup, including data collection, baselines, metrics, and hyperparameters, and then present the results. Our code is available at ~\cite{CAI_Learning_Backdoors_for_2024}.

\subsection{Problem Domains and Instance Generation}
We evaluate on several ILP problem domains, Generalized Independent Set Problem (GISP), Set Cover (SC), Combinatorial Auction (CA), and Maximum Independent Set (MIS) that are widely used in existing studies~\cite{song2020general,ferber2021learning,huang2023searching}. For each problem domain, we generate Small (S) and Large (L) instances for it. Following~\cite{ferber2021learning}, GISP-S and GISP-L instances are generated with 150 nodes and  175 nodes, respectively, with the node reward set to 100 and edge removal cost set to 1; In order to keep a similar evaluation procedure (e.g., Gurobi being able to solve to optimality within certain runtime) and diverse hardness of the problems, we choose the size of the three other problem domains as follows: Following \cite{huang2023searching}, SC-S instances have 1,000 variables and 1,200 constraints and SC-L instances have 1,000 variables and 1,500 constraints, with density set to 0.05 for both; Following ~\cite{leyton2000towards}, CA-S instances are generated with 150 items and 750 bids and CA-L instances are generated with 200 items and 1,000 bids according to the arbitrary relations; Following \cite{huang2023searching}, MIS-S and MIS-L instances are generated according to the Erdos-Renyi random graph model~\cite{erdHos1960evolution}, with an average degree of 4 and 1,250 and 1,500 nodes, respectively. 

To show our method can work on general MILP cases, we also evaluate on Facility Location (FC) and Neural Network Verification (NN). Following ~\cite{gasse2019exact}, FC instances are generated with 100 facilities supplying 200 customers. Following ~\cite{nair2020solving}, NN instances are a convolutional neural network is verified on each image in the MNIST dataset, giving rise to a corresponding dataset of MILPs. Table \ref{table:a} shows average numbers of variables, constraints, and average Gurobi solving time of 100 test instances from each problem domain. More details of instance generation are included in the Appendix~\cite{CAI_Learning_Backdoors_for_2024}.

We refrained from evaluating our methodology on heterogeneous MILP datasets like MIPLIB~\cite{gleixner2021miplib} due to the challenges in training an effective model using backdoors collected from a variety of MILP distributions. Instead, we target homogeneous settings from Distributional MIPLIB~\cite{huang2024distributional} because many real-world scenarios require solving a homogeneous family of problems, where instances share similar structures but differ slightly in problem size or numerical coefficients. Since MCTS backdoor collection in~\cite{khalil2022finding} only works with mixed-binary instances due to the ease of implementation of the tree weight for binary problems, we have limited our problem domains to MILP with only binary variables. In practice, we believe contrastive learning-based training still applies when using LP-based sampling to collect data. 

\vspace{10pt}
\subsection{Experiment Setup}
\begin{table*} [ht]
\centering
\caption{Runtime (secs) of Gurobi (\textsc{grb}), the scorer model (\textsc{scorer}), the scorer with the classifier (\textsc{scorer}+\textsc{cls}), the contrastive learning model (\textsc{cl}) and the contrastive learning model with the classifier (\textsc{cl} + \textsc{cls}) on GISP-S, SC-S, CA-S and MIS-S. Scorer and \textsc{cl} are trained on 200 instances and the classifiers are trained on another 200 instances. We report the mean, the standard deviation, 25th, 50th, and 75th percentiles, and win/tie/loss (W/T/L) counts vs. Gurobi. We report the speed improvement as a percentage inside the parentheses for the mean and median. We bold the best-performing entries.\\}
\begin{tabular}{llllllll}
\toprule
Dataset (\#vars, \#cons) & Solver & Mean & Std Dev & 25 pct & Median & 75 pct & W/T/L vs \textsc{grb} \\
\midrule
\multirow{5}{*}{GISP-S (988, 3253)} & \textsc{grb} & 633 & 154 & 513 & 615 & 731 & -\\
& \textsc{scorer} & 544 (14.1\%) & \textbf{117} & 453 & 543 (11.7\%) & \textbf{611} & 74/0/26\\ 
& \textsc{scorer} + \textsc{cls} & 578 (8.7\%) & 128 & 502 & 586 (4.7\%) & 637 & 43/32/25\\
& \textbf{\textsc{cl}} & \textbf{533 (15.8\%)} & 120 & \textbf{441} & \textbf{524 (14.8\%)} & 612 & \textbf{84/0/16}\\
& \textsc{cl} + \textsc{cls} & 560 (11.5\%) & 135 & 462 & 555 (9.8\%) & 653 & 59/27/14\\

\midrule
\multirow{5}{*}{SC-S (1000, 1200)} & \textsc{grb} & 171 & 189 & 50.8 & 101 & 216 & - \\
& \textsc{scorer} & 244 (-42.7\%) & 219 & 67.2 & 148 (-46.5\%) & 295 & 27/0/73 \\
& \textsc{scorer} + \textsc{cls} & 203 (-18.7\%) & 195 & 62.1 & 142 (-40.6\%) & 256 & 15/57/28\\
& \textbf{\textsc{cl}} & \textbf{149 (12.9\%)} & \textbf{163} & \textbf{42.5} & 79.8 (20.1\%) & \textbf{195} & \textbf{77/0/23}\\
& \textsc{cl} + \textsc{cls} & 153 (10.5\%) & 165 & 49.4 & \textbf{79.4 (21.4\%)} & 197 & 52/27/21\\
\midrule
\multirow{5}{*}{CA-S (750,282)} & \textsc{grb} & 177 & 94.6 & 110 & 151 & 229 & - \\
& \textsc{scorer} & 224 (-26.6\%) & 120 & 125 & 217 (-43.7\%) & 293 & 17/0/83 \\
& \textsc{scorer} + \textsc{cls} & 194 (-9.6\%) & 112 & 119 & 191 (-26.5\%) & 243 & 14/52/34\\
& \textbf{\textsc{cl}} & \textbf{156 (11.9\%)} & \textbf{83.0} & \textbf{88.5} & 146 (3.3\%) & \textbf{195} & \textbf{68/0/32}\\
& \textsc{cl} + \textsc{cls} & 170 (4.0\%) & 96.2 & 103 & \textbf{139 (8.0\%)} & 227 & 46/28/26\\

\midrule
\multirow{5}{*}{MIS-S (1250, 3946)} & \textsc{grb} & 218 & 231 & 79.7 & 147 & 267 & - \\
& \textsc{scorer} & 236 (-8.3\%) & 258 & 89.9 & 156 (-6.1\%) & 269 & 36/0/64\\
& \textsc{scorer} + \textsc{cls} & 219 (-0.5\%) & 237 & 82.3 & 154 (-4.8\%) & 264 & 24/46/30\\
& \textbf{\textsc{cl}} & \textbf{184 (15.6\%)} & \textbf{179} & \textbf{76.8} & \textbf{127 (13.6\%)} & \textbf{214} & \textbf{69/0/31}\\
& \textsc{cl} + \textsc{cls} & 190 (12.8\%) & 197 & 80.2 & 136 (7.5\%) & 227 & 44/35/21\\
\bottomrule
\end{tabular}%
\label{table:2}
\end{table*}
\vspace{10pt}
\begin{table*}[ht]
\centering
\caption{Runtime (secs) of Gurobi (\textsc{grb}), the scorer with the classifier (\textsc{scorer}+\textsc{cls}), the contrastive learning model (\textsc{cl}) on FC and NN. We report the mean, the standard deviation, 25th, 50th, and 75th percentiles, and win/tie/loss (W/T/L) counts vs. Gurobi. We report the speed improvement as a percentage inside the parentheses for the mean and median. We bold the best-performing entries.\\}
\begin{tabular}{llllllll}
\toprule
Dataset (\#binvars, \#cons) & Solver & Mean & Std Dev & 25 pct & Median & 75 pct & W/T/L vs \textsc{grb} \\
\midrule
\multirow{3}{*}{FC (100, 40000)} & \textsc{grb} & 46.7 & 32.9 & \textbf{20.6} & 38.0 & 59.7 & -\\
& \textsc{scorer+cls} & 44.3 (5.1\%) & 30.2 & 25.1 & 35.5 (6.6\%) & 53.1 & 28/65/7 \\ 
& \textbf{\textsc{cl}} & \textbf{39.6 (15.2\%)} & \textbf{20.2} & 25.2 & \textbf{33.4 (12.1\%)} & \textbf{46.9} & \textbf{60/0/40} \\
\midrule
\multirow{3}{*}{NN (167, 6520)} & \textsc{grb} & 9.24 & 20.4 & 3.46 & 4.83 & 7.94 & -\\
& \textsc{scorer+cls} & 8.14 (11.9\%) & 19.5 & 3.16 & 4.26 (11.8\%) & 6.95 & 62/21/17 \\ 
& \textbf{\textsc{cl}} & \textbf{7.81 (15.5\%)} & \textbf{15.6} & \textbf{3.12} & \textbf{4.13 (14.5\%)} & \textbf{6.75} & \textbf{80/0/20} \\
\bottomrule
\end{tabular}
\label{table:5}
\end{table*}

\textbf{Data Collection} \space\space We generate 200 small instances for each problem domain. We use the MCTS framework to collect backdoors of size 8 for each instance over 5 hours with 10 parallel workers. We follow~\cite{khalil2022finding} and choose 8, which shows promising results on finding high-quality backdoors on 164 problem domains From MILPLIB2017. We select the top 50 backdoors measured by tree weight function, solve the problems using those backdoors with MILP solver, and record the runtime. For contrastive loss, we select the 5 fastest candidate backdoors as positive samples and 5 slowest candidate backdoors as negative samples.
\\
\\
\noindent \textbf{Baselines} \space\space We compare our contrastive learning model (\textsc{cl}) with several baselines:  Gurobi (\textsc{grb}), the scorer model (\textsc{scorer}) and the scorer with the classifier (\textsc{scorer}+\textsc{cls}). \textsc{scorer} is the previous method ~\cite{ferber2021learning} that predicts the score for candidate backdoors using supervised learning and \textsc{scorer}+\textsc{cls} is \textsc{scorer} with \textsc{cls} to predict whether to use the backdoor output by the score or not. The \textsc{scorer} and \textsc{cl} models are trained with 200 instances, with 160 and 40 instances for training and validation, respectively. For \textsc{cls}, we generate another 200 instances, with 160 and 40 instances for training and validation, respectively. 
%To train the classifier model, we generate the dataset with the help of the scorer or \textsc{cl} model we trained previously and need to complete only one backdoor data collection with very short time compare to MCTS. 
To train the classifier for \textsc{scorer}+\textsc{cls}, we use random sampling based on the LP relaxation \cite{dilkina2009backdoors} to generate random 50 candidate backdoors and use \textsc{scorer} to select the best one to collect runtime statistics. Then, we assign labels for each instance based on the runtime of the selected backdoor over the Gurobi runtime and use a cross-entropy loss to train the classifier following~\cite{ferber2021learning}. 
To train the classifier for \textsc{cl}+\textsc{cls}, we collect the runtime statistics from the backdoor output directly from \textsc{cl} model and the rest is the same as \textsc{scorer}+\textsc{cls}. %instead of choosing from a candidate sets and generate the dataset for training the classifier the same as \textsc{scorer}. 

We did not compare other learning for Branch-and-Bound ~\cite{song2018learning, gasse2019exact} baselines because we believe their works are conjunctive of ours. Our model just predicts a small set of variables before the solving, and during the solving; it is still applicable to use other decision-making approaches. Also, their approaches required the specific implementation of solvers, which limited them to open-source solver SCIP, while our approach can be applied to the current state-of-art solver Gurobi, which makes our baseline more competitive than theirs.
\\
\\
\noindent \textbf{Metrics} \space\space To compare the performance of the methods, we test on 100 instances for each problem domain and report summary statistics of runtime to solve the problem to optimality, including mean, standard deviation, 25th percentile, median, and 75th percentile. We also report the wins, ties, and losses over Gurobi. We consider the runtime of the full pipeline, including feature extractor, inference for the GAT model, sampling or greedy selection, and solving the MILP. When \textsc{cls} suggests using Gurobi, we record a “tie” to give insight into model predictions even though there is a slight loss in runtime due to the ML overhead.
\\
\\
\begin{table*}[ht]
\centering
\caption{Evaluating generalization to larger problem size: Runtime (secs) of standard Gurobi (\textsc{grb}) and the contrastive learning model (\textsc{cl}) on GISP-L, SC-L, CA-L, MIS-L. The model is trained on 200 small instances for each problem domain. We report the mean, the standard deviation, 25th, 50th, and 75th percentiles, and win/loss (W/L) counts vs. Gurobi. We report the speed improvement as a percentage inside the parentheses for the mean and median. We bold the best-performing entries.\\}
\begin{tabular}{llllllll}
\toprule
Dataset (\#vars, \#cons) & Solver & Mean & Std Dev & 25 pct & Median & 75 pct & W/L vs \textsc{grb} \\
\midrule
\multirow{2}{*}{GISP-L (1316, 4571)} & \textsc{grb} & 3,363 & 989 & 2,611 & 3,303 & 4,049 & -\\
& \textbf{\textsc{cl}} & \textbf{2,879 (14.4\%)} & \textbf{785} & \textbf{2,288} & \textbf{2,910 (11.9\%)} & \textbf{3,331} & \textbf{77/23}\\
\midrule
\multirow{2}{*}{SC-L (1000, 1500)} & \textsc{grb} & 1,195 & 1,692 & 161 & 519 & 1,597 & - \\
& \textbf{\textsc{cl}} & \textbf{1,003 (16.1\%)} & \textbf{1,433} & \textbf{143} & \textbf{435 (16.0\%)} & \textbf{1,309} & \textbf{78/22}\\
\midrule
\multirow{2}{*}{CA-L (1000, 377)} & \textsc{grb} & 1,213 & 625 & 881 & 1,093 & 1,353 & - \\
& \textbf{\textsc{cl}} & \textbf{1,096 (9.6\%)} & \textbf{487} & \textbf{767} & \textbf{994 (9.1\%)} & \textbf{1,159} & \textbf{69/31}\\
\midrule
\multirow{2}{*}{MIS-L (1500, 5941)} & \textsc{grb} & 759 & 1,356 & 128 & 263 & 1,023 & - \\
& \textbf{\textsc{cl}} & \textbf{550 (27.5\%)} & \textbf{987} & \textbf{127} & \textbf{256 (2.7\%)} & \textbf{501} & \textbf{75/25}\\
\bottomrule
\end{tabular}%
\label{table:3}
\end{table*}

\begin{figure*}[ht]
 \begin{subfigure}{0.49\textwidth}
     \includegraphics[width=\textwidth]{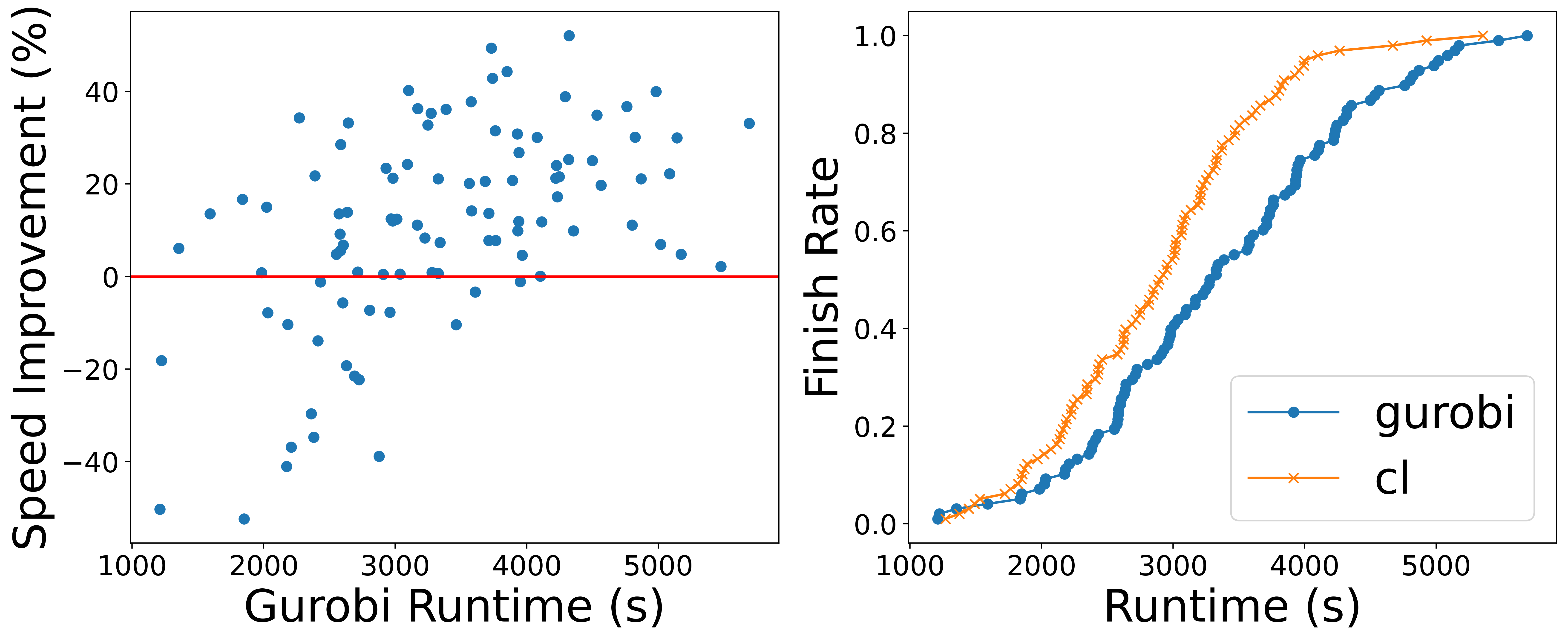}
     \caption{GISP-L}
     \label{fig:a}
 \end{subfigure}
 \hfill
 \begin{subfigure}{0.49\textwidth}
     \includegraphics[width=\textwidth]{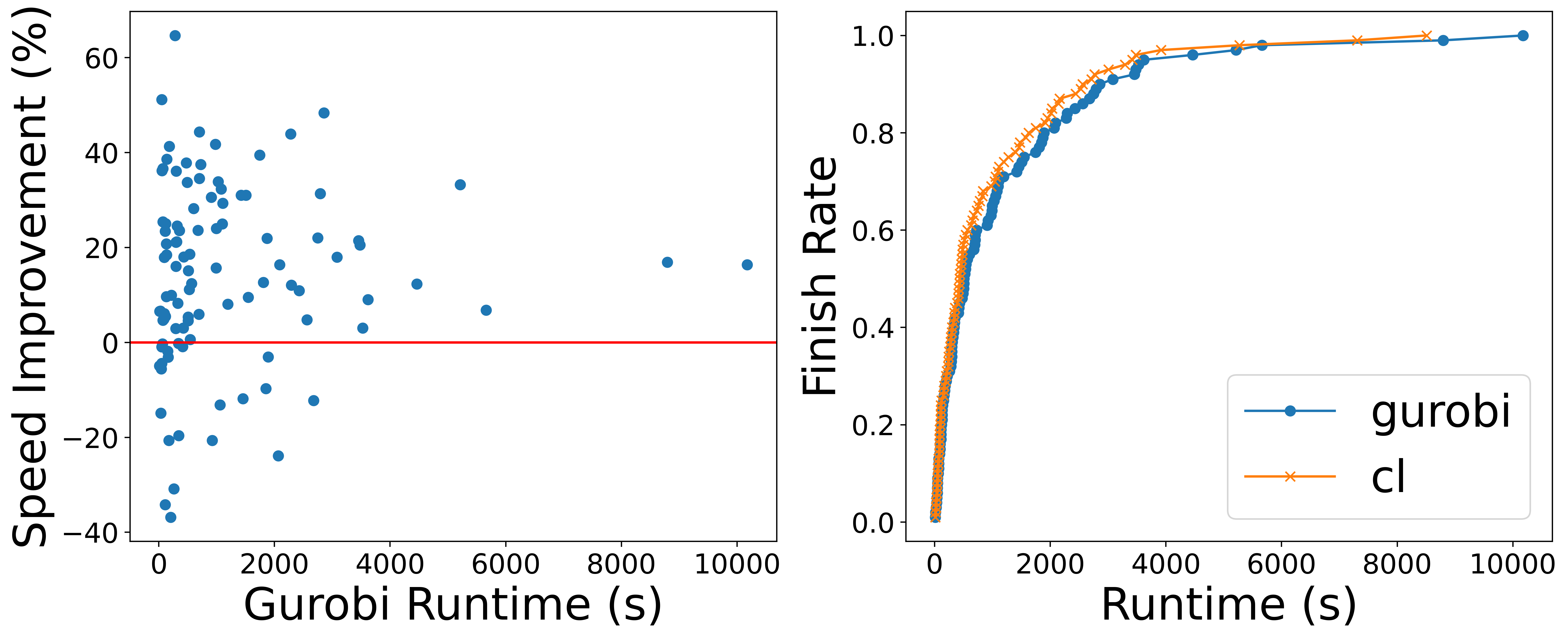}
     \caption{SC-L}
     \label{fig:b}
 \end{subfigure}
\\
\\
 \begin{subfigure}{0.49\textwidth}
     \includegraphics[width=\textwidth]{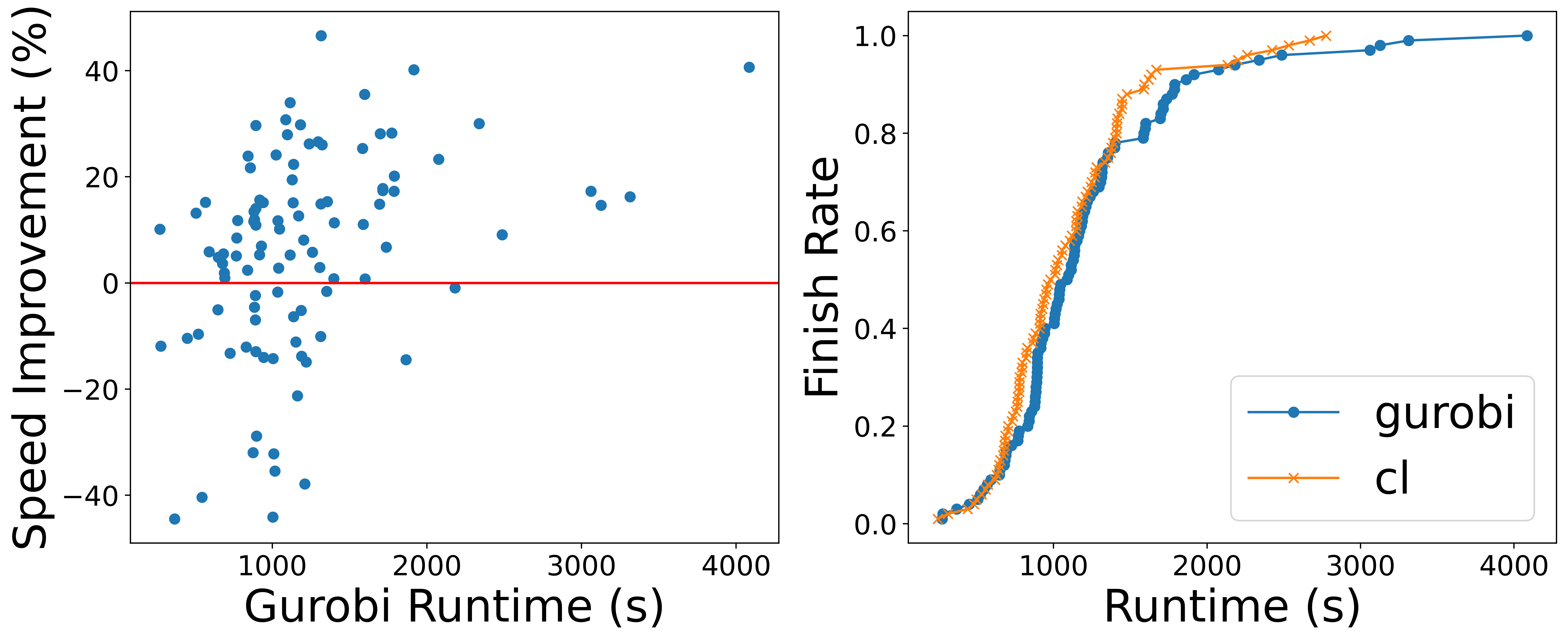}
     \caption{CA-L}
     \label{fig:c}
 \end{subfigure}
 \hfill
 \begin{subfigure}{0.49\textwidth}
     \includegraphics[width=\textwidth]{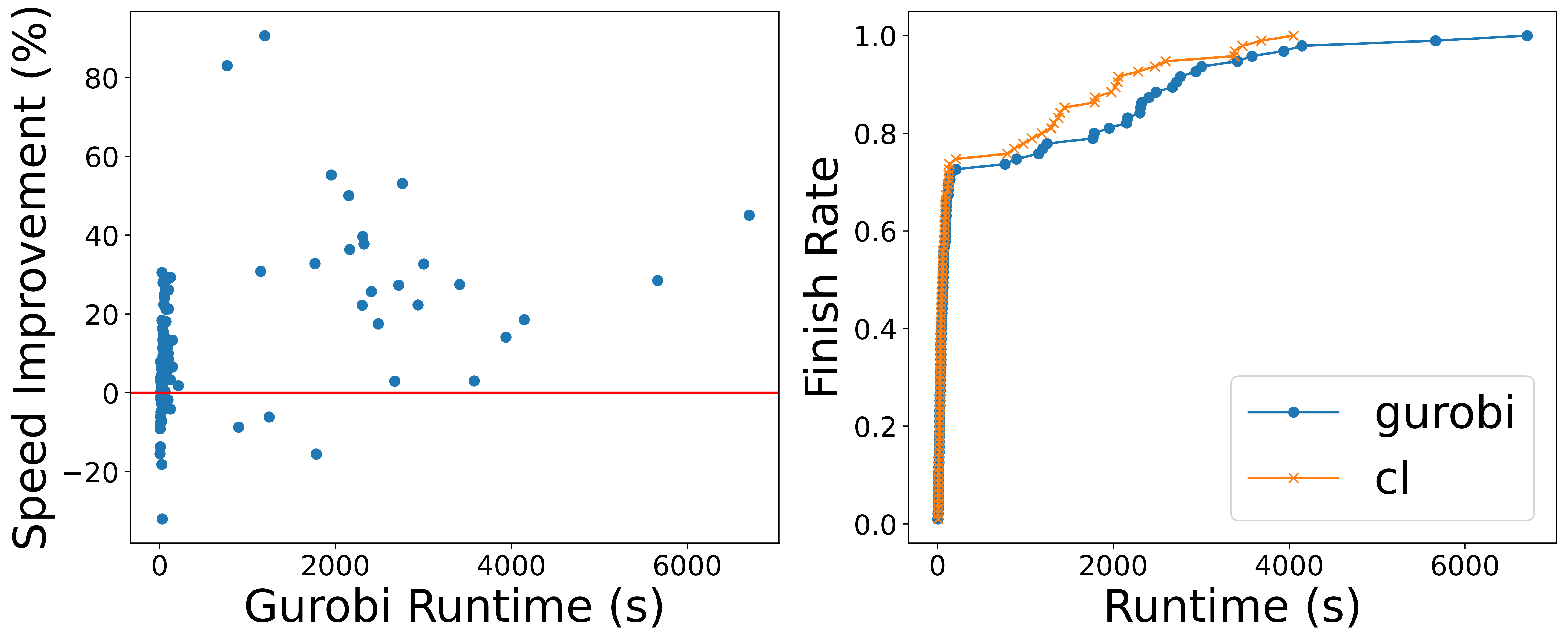}
     \caption{MIS-L}
     \label{fig:d}
 \end{subfigure}
 \\
 \caption{This figure shows the runtime of Gurobi (\textsc{grb}) and contrastive learning model (\textsc{cl}) on GISP-L, SC-L, CA-L, and MIS-L through two different types of plots. The left part is a scatter plot with Gurobi runtime as the x-axis and the speed improvement as the percentage of \textsc{cl} over \textsc{grb} as the y-axis. The points above the red line are ones where \textsc{cl} is better than \textsc{grb} and vice versa. The right part shows the finish rate as a function of runtime. The finish rate for a given runtime is the fraction of instances solved to optimality within the runtime. Every dot on the lines indicates one finished instance. The figures show that \textsc{cl} outperforms \textsc{grb} on average and specifically provides speedups on the harder instances in each distribution.\\}
 \label{fig:2}
\end{figure*}

\noindent \textbf{Hyperparameters} \space\space We solve MILPs with single-threaded Gurobi 10.0 on four 64-core machines with Intel 2.1GHz CPUs and 264GB of memory. Training is done on an NVIDIA Tesla V100 GPU with 112GB memory. For \textsc{cl} model, we use the Adam optimizer~\cite{kingma2014adam} with a learning rate of $5\times 10^{-4}$ and a weight decay of 0.01. We use a batch size of 32 and train for 100 epochs. For \textsc{scorer} and \textsc{cls}, the learning rate is set to $5\times 10^{-3}$ and a batch size of 128 and train for 100 epochs. 
Due to the fact of pairwise ranking loss in the \textsc{scorer}, the training for \textsc{scorer} is extremely expensive, which takes around 10-15 hours, while the training time for \textsc{cl} model is less than 30 minutes. The more efficient use of training data significantly improves our \textsc{cl} model. 
\\
\\
\noindent \textbf{Experiments} \space\space We conduct the following sets of experiments:
\begin{itemize}
\item We compare \textsc{cl} and \textsc{cl}+\textsc{cls} against \textsc{grb},  \textsc{scorer}, \textsc{scorer}+\textsc{cls} on small ILP domains; 
\item We compare \textsc{cl} against \textsc{grb}, \textsc{scorer}+\textsc{cls} on MILP domains.
\item We test the generalizability of the \textsc{cl} model (trained on the small instances) on large ILP domains and compare it against \textsc{grb}.
\item Ablation study: For GISP-S, SC-S, we test how contrastive learning and MCTS data collection separately affect the performance of backdoor predicting.
\end{itemize}

\begin{table*}
\caption{Evaluating the role of the backdoor collection strategy: Runtime (secs) of standard Gurobi (\textsc{grb}), the \textsc{scorer} model (\textsc{scorer}), and the contrastive learning model (\textsc{cl}) with different backdoor collection methods, MCTS, and sampling. We report the mean, the standard deviation, 25th, 50th, and 75th percentiles, and win/loss (W/L) counts vs. Gurobi. We report the speed improvement as a percentage inside the parentheses for the mean and median. We bold the best-performing entries.\\}
\centering
\begin{tabular}{llllllll}
\toprule
Dataset (\#vars, \#cons) & Solver & Mean & Std Dev & 25 pct & Median & 75 pct & W/L vs \textsc{grb} \\
\midrule
\multirow{5}{*}{GISP-S (988, 3253)} 
& \textsc{grb} & 633 & 154 & 513 & 615 & 731 & - \\
& \textsc{scorer} + Sampling & 601 (5.0\%) & 152 & 518 & 580 (5.7\%) & 681 & 49/51\\
& \textsc{scorer} + MCTS & 544 (14.1\%) & \textbf{117} & 453 & 543 (11.7\%) & \textbf{611} & 74/26\\ 
&\textsc{cl} + Sampling & 565 (10.7\%) & 132 & 455 & 556 (9.6\%) & 634 & 67/33\\
& \textbf{\textsc{cl} + MCTS} & \textbf{533 (15.8\%)} & 120 &\textbf{441} &\textbf{524 (14.8\%)} & 612  & \textbf{84/16} \\
\midrule
\multirow{5}{*}{SC-S (1000, 1200)} 
& \textsc{grb} & 171 & 189 & 50.8 & 101 & 216 & -  \\
& \textsc{scorer} + Sampling & 287 (-67.8\%) & 358 & 85.4 & 167 (-65.3\%) & 359 & 3/97\\
& \textsc{scorer} + MCTS & 244 (-42.7\%) & 219 & 67.2 & 148 (-46.5\%) & 295 & 27/73 \\
& \textsc{cl} + Sampling & 186 (-8.8\%) & 196 & 48.5 & 107 (-5.9\%) & 254 & 43/57\\
& \textbf{\textsc{cl} + MCTS} & \textbf{149 (12.9\%)} & \textbf{163} & \textbf{42.5} & \textbf{79.8 (20.1\%)} & \textbf{195} & \textbf{77/23} \\
\bottomrule

\end{tabular}%
\label{tab:my_label}
\end{table*}

\subsection{Results}
\textbf{Backdoor Collection Quality} \space\space For each problem domain, we gathered solve times from 50 MCTS selected candidate backdoors across 200 instances. This resulted in a dataset comprising solve times for 10,000 candidate backdoors. Figure \ref{fig:bd} displays histograms of the normalized runtimes, which we calculated by dividing the solve time with a candidate backdoor by the original solve time for each of the 10,000 candidate backdoors in every problem domain. We observe significant improvement in MCTS data collection over the previous sampling method. Specifically, we noted that over half of the candidate backdoors outperformed the baseline solve time in the GISP-S, FC, and NN problem domains. In stark contrast, the sampling method employed in ~\cite{ferber2021learning} failed to identify any backdoors that improved the original solve times for SC-S, CA-S, and MIS-S domains. However, our approach demonstrated a considerable portion of candidate backdoors with better performance. The improvement of candidate backdoors directly leads to a higher quality of training data, which helps train more accurate models for backdoor prediction.
\\

\noindent \textbf{ILP domains performance} \space\space  We present results on 100 small test instances in Table \ref{table:2}. \textsc{cl} consistently performs better than \textsc{scorer} and has more than 10\% improvement on the average runtime over \textsc{grb} in all problem domains, even in those where the \textsc{scorer} does not work. The \textsc{scorer} performs well on GISP-S with 14.1\% speed improvement but performs worse than \textsc{grb} on SC-S, CA-S, MIS-S with 67.8\%, 26.6\%, 8.3\% increase on the average runtime. When the problem domains become more varied and less structured, the \textsc{scorer} stopped functioning due to issues inherent in the sampling process. However, in \textsc{cl}, we benefit from greedy selection, and as the experiment results show, the backdoor output by \textsc{cl} has high quality, leading to 77, 68, 69 wins over Gurobi on SC-S, CA-S, and MIS-S, respectively. We can see \textsc{cls} indeed helps the \textsc{scorer} in most domains to improve the average performance, but \textsc{cls} does not fit well with our \textsc{cl}, dropping the performance range from 2.4\% to 7.9\%. Consider a real-world problem domain with a limited number of instances, using all the data to train \textsc{cl} would be a better and more elegant method.
\\
\\
\noindent \textbf{MILP domains performance} \space\space Following ~\cite{ferber2021learning}, we present results on 100 test instances from FC and NN in Table \ref{table:5} to show our method can work in general MILP settings. \textsc{cl} still performs better than \textsc{grb} and \textsc{scorer+cls} in both domains, with around 15\% improvement over \textsc{grb} and 10\% improvement over \textsc{scorer+cls} on average runtime. Although \textsc{scorer+cls} have fewer losses compared to \textsc{grb} by transforming them into ties, \textsc{cl} have great improvement on win rate over \textsc{scorer+cls}, from 28 to 60 in FC and from 62 to 80 in NN.
\\
\\
\noindent \textbf{Generalization performance} \space\space We test \textsc{cl} trained on 100 small instances on 100 large instances of the same problem domain. This experiment shows that when the instances are large and MCTS becomes too inefficient to collect data, we can still use \textsc{cl} trained on smaller instances. As the results shown in Table \ref{table:3}, \textsc{cl} still consistently outperforms \textsc{grb}. On GISP-L and CA-L, \textsc{cl} achieves only a slight percentage drop on mean over \textsc{grb} compared to the small instances; On SC-L and MIS-L even achieve better performance. The results show that \textsc{cl} also performs well on larger and harder instances, as illustrated by Figure \ref{fig:2}. The figure compares the runtime of \textsc{grb} and \textsc{cl} in terms of speed improvement and finish rate. On the left part, the dots under the red line are mostly easy instances, and our model almost gains 100\% win for the hard instances. On the right side, the orange line is consistently above the blue line, except for the starting line, which is an easy instance. The results clearly show \textsc{cl} has great generalizability to the problem with the same domain and opens the potential for learning backdoors on much larger real-world problem domains. 
\\
\\
\noindent \textbf{Ablation Study} \space \space We evaluate how contrastive learning and the MCTS backdoor collection contribute to the model performance. We collect another set of backdoor data on the same instances using LP-based sampling, the backdoor collection method used by~\cite{ferber2021learning}. We compare the performance of \textsc{scorer} + Sampling, \textsc{scorer} + MCTS, \textsc{cl} +Sampling, and \textsc{cl} + MCTS on GISP-S and SC-S and Table \ref{tab:my_label} reports the runtime statistics. Overall, \textsc{cl} + MCTS performs best on GISP-S and SC-S. The results show performance improvement for both \textsc{scorer} and \textsc{cl} when using MCTS, demonstrating the benefit of replacing random sampling with MCTS for data collection. The results also show that the \textsc{cl} consistently performs better than the \textsc{scorer} with the same data collection method, showing the benefit of contrastive learning over the previous learning-to-rank method. Additionally, the performance improvement in \textsc{cl} + Sampling shows the potential of our contrastive learning when trained with LP-based sampling backdoors to be applied to MILP domains with non-binary integers. 

\section{Conclusion}
This paper introduces a contrastive learning framework to predict effective backdoors for MILPs. We utilize a Monte-Carlo tree search method to collect training data of better quality than previous methods. Empirical results on several common problem domains indicated that the contrastive learning model significantly outperforms existing baselines. It also achieved good generalization performance on larger-size instances. Beyond finding better backdoors, our model requires less time and resources for training and reduces randomness compared to the previous scorer model. The limitation is that the model requires offline data collection and training for each problem distribution to reach good accuracy. %With the better model and better backdoor data collection, we are eager to find real-world problem domains with backdoor quality, which is a very simple add-on to solving the problem but can improve performance. 
For future work, we want to develop some theory related to backdoors to understand the theoretical properties of backdoors. Moreover, given the recent successes of contrastive learning, it can be applied to other MILP learning problems, potentially becoming a foundational method for ML in combinatorial optimization.

\section*{Acnowledgement}
The research was supported by the National Science Foundation (NSF) under grant number 2112533.

%%%%%%%%%%%%%%%%%%%%%%%%%%%%%%%%%%%%%%%%%%%%%%%%%%%%%%%%%%%%%%%%%%%%%%%%

%%% Use this command to include your bibliography file.

\bibliography{ijcai24}

\appendix
\section{Network Architecture}
We give full details of the GAT architecture described in Section 3.2. The policy takes the bipartite graph $P' = (G,V,C,E)$ as input and output a score vector with one score per decision variable. We use 2-layer MLPs with 64 hidden units per layer and ReLU as the activation function to map each node feature and edge feature to $\mathbb{R}^L$ where $L = 64$.

Let $\boldsymbol{v}_j, \boldsymbol{c}_i, \boldsymbol{e}_{i,j} \in \mathbb{R}^L$ be the embeddings of the $i$-th variable, $j$-th constraint and the edge connecting them output by the embedding layers $V^1, C^1, E^1$. We perform two rounds of message passing through the GAT. In the first round, each constraint node $\boldsymbol{c}_i$ attends to its neighbors $\mathcal{N}_i$ using an attention structure with $H = 8$ attention heads:
\begin{align*}
     \boldsymbol{c}'_{i} = \frac{1}{H}\sum_{i=1}^{H} \left(\alpha^{(h)}_{ii,1}\boldsymbol{\theta}^{(h)}_{c,1}\boldsymbol{c}_i + \sum_{j\in\mathcal{N}_i}\alpha^{(h)}_{ij,1}\boldsymbol{\theta}^{(h)}_{v,1}\boldsymbol{v}_j \right) 
\end{align*}
where $\boldsymbol{\theta}^{(h)}_{c,1} \in \mathbb{R}^{L\times L}$ and $\boldsymbol{\theta}^{(h)}_{v,1} \in \mathbb{R}^{L\times L}$ are learnable weights. The updated constraints embeddings $\boldsymbol{c}'_i$ in updated embedding layer $C^2$ are averaged across $H$ attention heads using attention weights
\begin{align*}
    \alpha^{(h)}_{ij,1} = \frac{\exp(\boldsymbol{w}^T_1\rho([\boldsymbol{\theta}^{(h)}_{c,1}\boldsymbol{c}_i, \boldsymbol{\theta}^{(h)}_{v,1}\boldsymbol{v}_j, \boldsymbol{\theta}^{(h)}_{e,1}\boldsymbol{e}_{i,j}]))}{\sum_{k\in \mathcal{N}_i}\exp(\boldsymbol{w}^T_1\rho([\boldsymbol{\theta}^{(h)}_{c,1}\boldsymbol{c}_i, \boldsymbol{\theta}^{(h)}_{v,1}\boldsymbol{v}_k, \boldsymbol{\theta}^{(h)}_{e,1}\boldsymbol{e}_{i,k}]))}
\end{align*}
where the attention coefficients $\boldsymbol{w}_1 \in \mathbb{R}^{3L}$ and $\boldsymbol{\theta}^{(h)}_{e,1} \in \mathbb{R}^{L\times L}$ are both learnable weights and $\rho(\cdot)$ refers to the LeakyReLU activation function with negative slope $0.2$. In the second round, similarly, each variable node attends to its neighbors to get updated variable node embeddings
\begin{align*}
     \boldsymbol{v}'_{j} = \frac{1}{H}\sum_{i=1}^{H} \left(\alpha^{(h)}_{jj,2}\boldsymbol{\theta}^{(h)}_{c,2}\boldsymbol{c}'_i + \sum_{j\in\mathcal{N}_i}\alpha^{(h)}_{ji,2}\boldsymbol{\theta}^{(h)}_{v,2}\boldsymbol{v}_j \right) 
\end{align*}
with attention weights
\begin{align*}
    \alpha^{(h)}_{ji,2} = \frac{\exp(\boldsymbol{w}^T_2\rho([\boldsymbol{\theta}^{(h)}_{c,2}\boldsymbol{c}'_i, \boldsymbol{\theta}^{(h)}_{v,2}\boldsymbol{v}_j, \boldsymbol{\theta}^{(h)}_{e,2}\boldsymbol{e}_{i,j}]))}{\sum_{k\in \mathcal{N}_i}\exp(\boldsymbol{w}^T_2\rho([\boldsymbol{\theta}^{(h)}_{c,2}\boldsymbol{c}'_i, \boldsymbol{\theta}^{(h)}_{v,2}\boldsymbol{v}_j, \boldsymbol{\theta}^{(h)}_{e,2}\boldsymbol{e}_{i,k}]))}
\end{align*}

where $\boldsymbol{w}_2 \in \mathbb{R}^{3L}$ and $\boldsymbol{\theta}^{(h)}_{c,2}, \boldsymbol{\theta}^{(h)}_{v,2}, \boldsymbol{\theta}^{(h)}_{e,2} \in \mathbb{R}^{L\times L}$ are learnable weights. After the two rounds of message passing, the final representations of variables $\boldsymbol{v}'$ in updated embedding layer $V^2$ are passed through a 2-layer MLP with 64 hidden units per layer to obtain a scalar value for each variable. Finally, we apply the sigmoid function to get a score between 0 and 1.

\section{Additional Details of Instance Generation}
We present the MILP formulations for the Generalized Independent Set Problem (GISP), Set Cover Problem (SC), Combinatorial Auction (CA), Maximal Independent Set (MIS), Facility Location (FC), and Neural Network Verification (NN).
\subsection{GISP}
In a GISP instance, we are given a graph $G$ with vertex set $V$ and edge set where $E_1$ is a set of non-removable edges and $E_2$ is a set of removable edges. A revenue $w_i > 0$ is associated with each vertex $i \in V$, and a cost $c_{ij} > 0$ is associated with each removable edge $(i, j) \in E_2$. The goal is to find an independent set, i.e., a set of vertices such that no two vertices in the set are adjacent, that maximizes the difference between the total revenue associated with the vertices in the set and the total cost associated with the removal of edges with both endpoints in the set:
\begin{gather*}
    \max \sum_{i\in V} w_ix_i - \sum_{(i,j)\in E_2}c_{ij}y_{ij} \\
    \text{s.t.  } x_i + x_j \leq 1, \forall (i,j) \in E_1, \\
    x_i + x_j -y_{ij} \leq 1, \forall (i,j) \in E_2, \\
    x_i \in \{0,1\}, \forall i \in V, \\
    y_{ij} \in \{0,1\}, \forall (i,j) \in E_2.
\end{gather*}

\subsection{SC}
In a SC instance, we are given $m$ elements and a collection $S$ of $n$ sets whose union is the set of all elements. The goal is to select a minimum number of sets from $S$ such that the union of the selected set is still the set of all elements:
\begin{gather*}
    \max -\sum_{s\in S}x_s \\ 
    \text{s.t.} \sum_{s\in S:i\in s} x_s \geq 1, \forall i \in [m], \\
    x_s\in\{0,1\}, \forall s \in S.
\end{gather*}

\subsection{CA}
In a CA instance, we are given $n$ bids $\{(B_i,p_i) : i\in [n]\}$ for $m$
items, where $B_i$ is a subset of items and $p_i$ is its associated bidding price. The objective is to allocate items to bids such that the total revenue is maximized:
\begin{gather*}
    \max \sum_{i\in[n]}p_ix_i \\
    \text{s.t.} \sum_{i:j\in B_i} x_i \leq 1, \forall j \in [m], \\
    x_i \in \{0,1\}, \forall i \in [n].
\end{gather*}

\subsection{MIS}
In a MIS instance, we are given an undirected graph $G = (V,E)$. The goal is to select the largest subset of nodes such that no two nodes in the subsets are connected by an edge in $G$:
\begin{gather*}
    \max \sum_{v\in V} x_v \\
    \text{s.t.} x_u + x_v \leq 1, \forall (u,v) \in E, \\
    x_v \in \{0,1\}, \forall v\in V.
\end{gather*}

\subsection{FC}
In a FC instance, we are given a set of potential facility locations $I$ and a set of customers $J$. Each facility $i \in I$ can serve multiple customers and has a fixed cost $f_i$ associated with opening it. Each customer $j \in J$ must be served by exactly one open facility, and the cost of serving customer $j$ from facility $i$ is denoted as $c_{ij}$. The goal is to minimize the total cost of opening facilities and serving all customers:
\begin{gather*}
    \min \sum_{i \in I} f_i y_i + \sum_{i \in I}\sum_{j \in J} c_{ij} x_{ij} \\
    \text{s.t.} \sum_{i \in I} x_{ij} = 1, \quad \forall j \in J, \\
    x_{ij} \leq y_i, \quad \forall i \in I, j \in J, \\
    \sum_{j \in J} D_j x_{ij} \leq S_i y_i, \quad \forall i \in I, \\
    y_i \in \{0,1\}, \quad \forall i \in I, \\
    x_{ij} \in \{0,1\}, \quad \forall i \in I, j \in J.
\end{gather*}
\vspace{-30pt}
\subsection{NN}
In a NN instance, we are given a neural network represented by a series of layers with nodes interconnected by weights. The goal is to verify whether, for all possible inputs within a specified range, the network's output adheres to certain safety or correctness criteria. The problem is typically formulated as determining if there exists any input that leads to an undesired or unsafe output, which can be expressed as:
\begin{gather*}
    \max \{z_k\} - c_k \quad  \\
    \text{s.t.} \quad x_{i+1} = f(W_i x_i + b_i), \quad \forall i , \\
    a \leq x_0 \leq b , \\
    z_k \text{ must satisfy safety criteria}, \\
    x_i, z_k \in \mathbb{R}^n .
\end{gather*}
\vspace{-30pt}
\section{Additional Details on Hyperparameter Tuning}
For \textsc{scorer} and \textsc{scorer+cls}, we use all the hyperparameters provided in Ferber et.al.'s code in our experiments.

We limit the number of backdoor samples $k$ to 50 for each instance during data collection due to the inefficiency of MCTS, thus giving us a restricted range of $p$ and $q$. We ran several experiments with minor, reasonable variations of $p$ and $q$ $(p=5, q=15)(p=10,q=10)$ during training, and we did not observe much difference in the performance of the resulting models. For $L$ and $H$, we are following previous work to avoid complicate ablation studies.

In our experiments, we greedily taking the $q$ worst candidate backdoors as negative samples. Other ways of determining negative samples have been experimented with, such as choosing the candidate backdoors with highest number of common variables but have worse performance or randomly sampling 5 backdoors from the 15 worst samples. The current strategy outperforms the others.

In Table \ref{fig:1}, we summarize all the hyperparameters with their notations and values used in our experiments.

\begin{table}[ht]
\centering
\caption{Hyperparameters with their notations and values used.}
\begin{tabular}{lcc}
\hline
\textbf{Hyperparameter} & \textbf{Notation} & \textbf{Value} \\
\hline
Number of candidate backdoors for each instance & $k$ & 50 \\
Number of positive samples & $p$ & 5 \\
Number of negative samples & $q$ & 5 \\
Feature embedding dimension & $L$ & 64 \\
Number of attention heads in the GAT & $H$ & 8 \\
Temperature parameter in the contrastive loss & $\tau$ & 0.07 \\
Learning rate  &  & $5 \times 10^{-4}$ \\
Weight decay & & 0.01 \\
Batch size  &  & 32 \\
Number of training epochs  &  & 100 \\
\hline
\end{tabular}
\label{fig:1}
\end{table}

\vspace{-10pt}
\section{Additional Figures}
Figure 1 presents a runtime comparison between the collected backdoors and the standard Gurobi runtime for GISP-S, SC-S, CA-S, MIS-S, FC, and NN. The histograms provide a visual representation, highlighting the instances where the best backdoor among the 50 collected outperforms the standard Gurobi runtime. Notably, GISP-S and FC exhibits the highest backdoor quality, followed by SC-S and NN, where the backdoor quality is considered good. However, for CA-S and MIS-S, the backdoor quality is less favorable, with a majority of backdoors unable to surpass the standard Gurobi runtime.

Figure 2 compares the runtime of \textsc{grb} and \textsc{cl} in terms of speed improvement and finish rate on small instances. On the left part, the dots under the red line are mostly easy instances, and our model almost gains most win for the hard instances. On the right side, the majority of the orange line is above the blue line. The results clearly show \textsc{cl} is better than \textsc{grb} in terms of running time.

\vspace{-10pt}
\begin{figure} [ht]
 \begin{subfigure}{0.24\textwidth}
     \includegraphics[width=\textwidth]{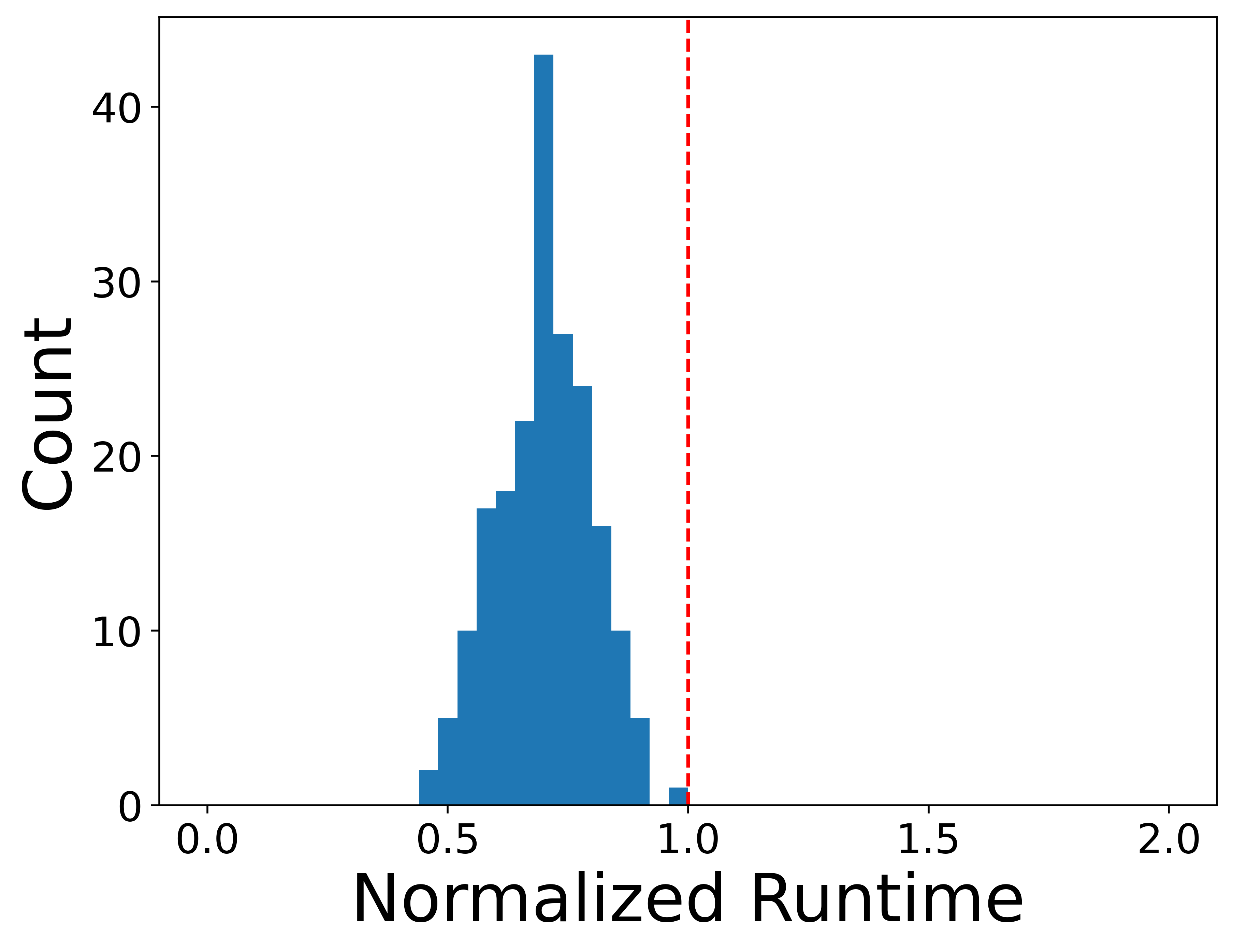}
     \caption{GISP-S}
     \label{fig:a}
 \end{subfigure}
 \begin{subfigure}{0.24\textwidth}
     \includegraphics[width=\textwidth]{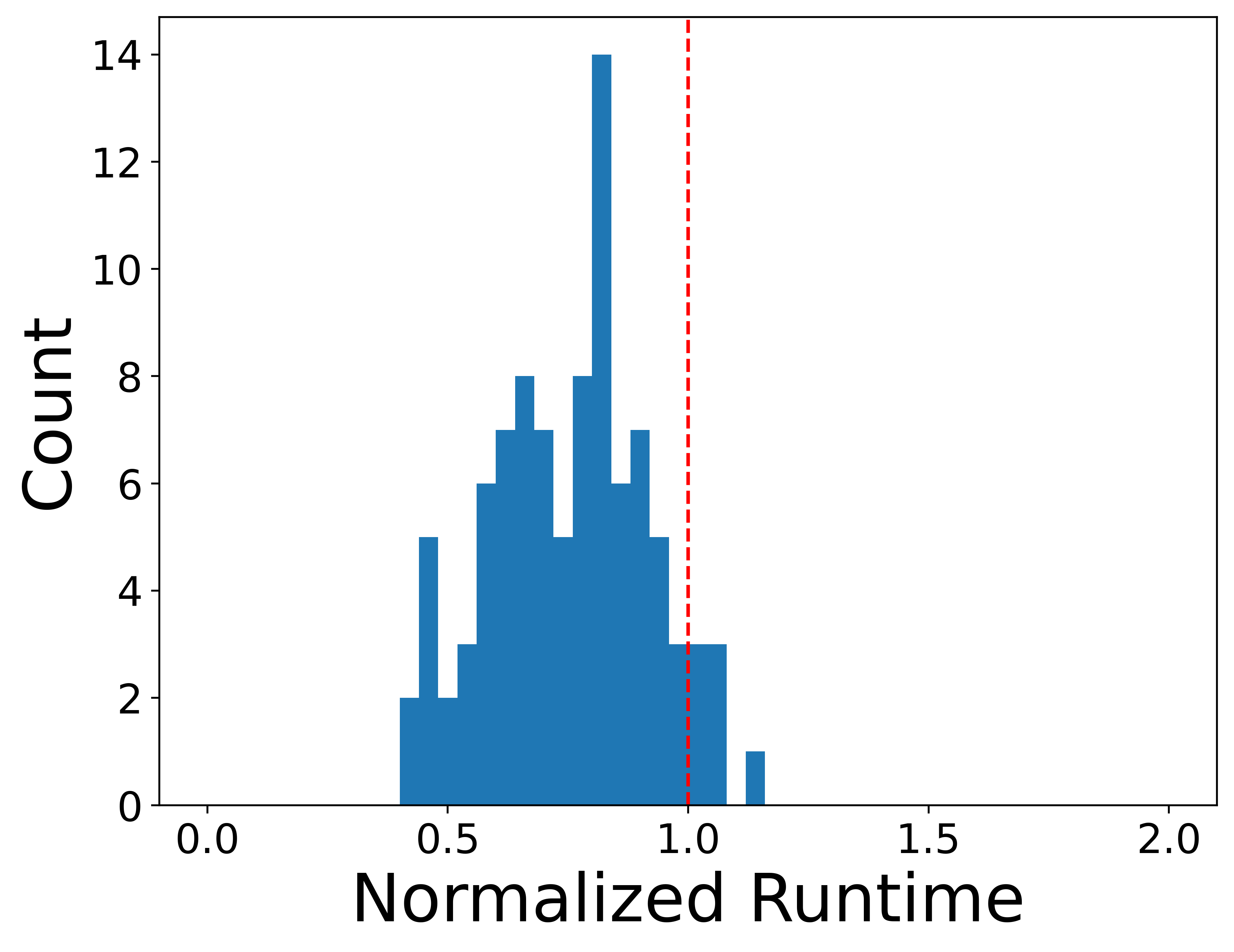}
     \caption{SC-S}
     \label{fig:b}
 \end{subfigure}
\\
\\
 \begin{subfigure}{0.24\textwidth}
     \includegraphics[width=\textwidth]{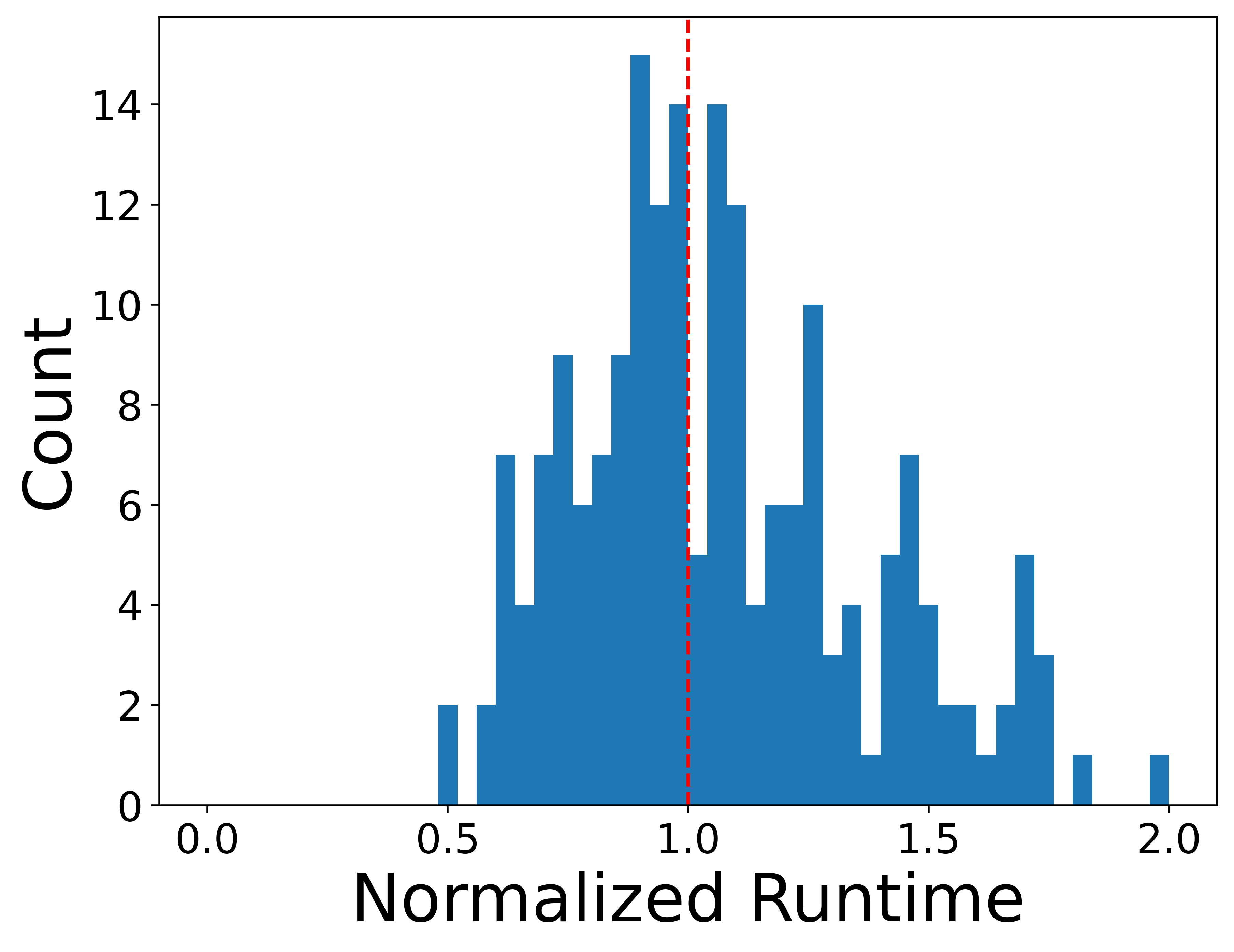}
     \caption{CA-S}
     \label{fig:c}
 \end{subfigure}
 \begin{subfigure}{0.24\textwidth}
     \includegraphics[width=\textwidth]{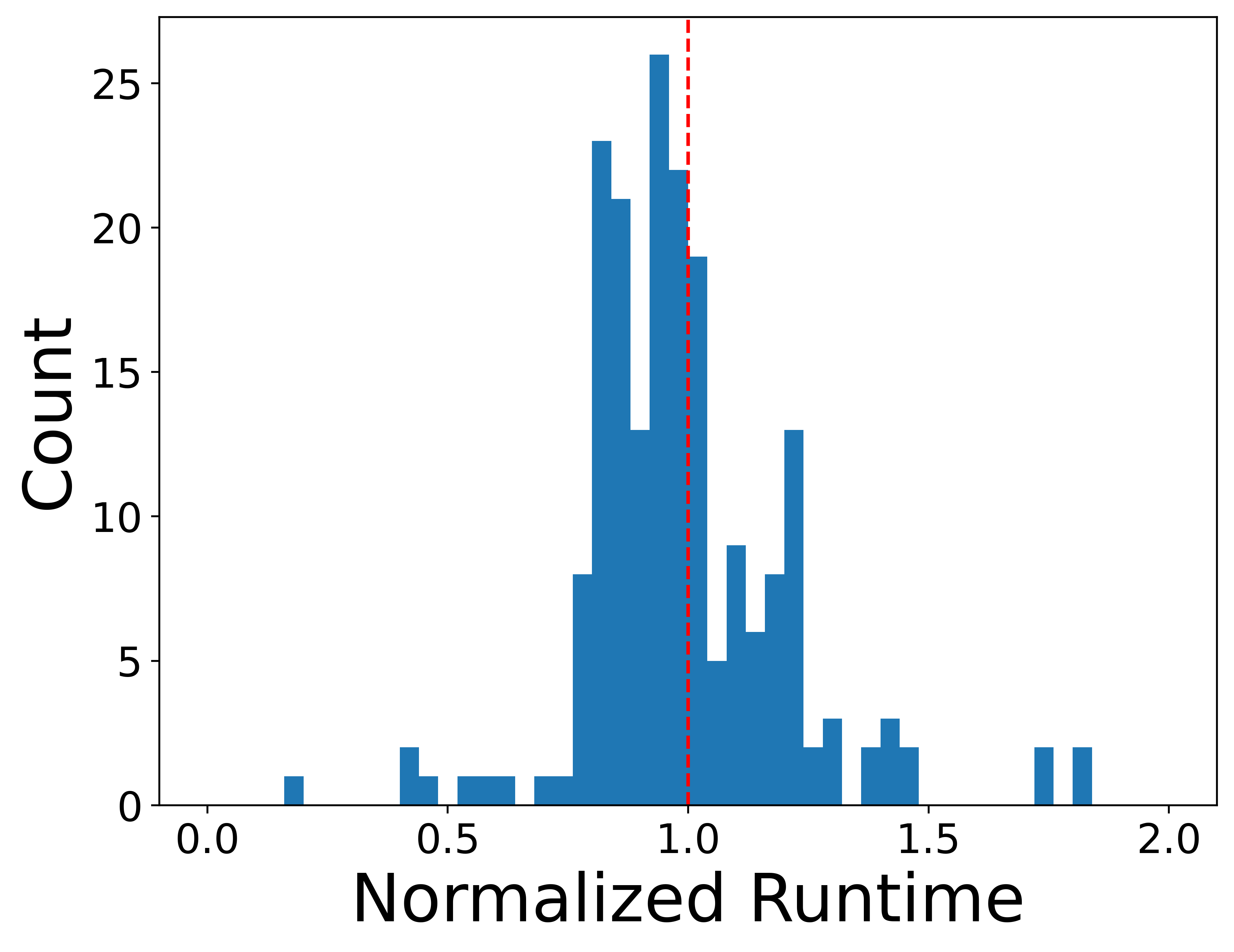}
     \caption{IS-S}
     \label{fig:c}
 \end{subfigure}
 \\
 \\
 \begin{subfigure}{0.24\textwidth}
     \includegraphics[width=\textwidth]{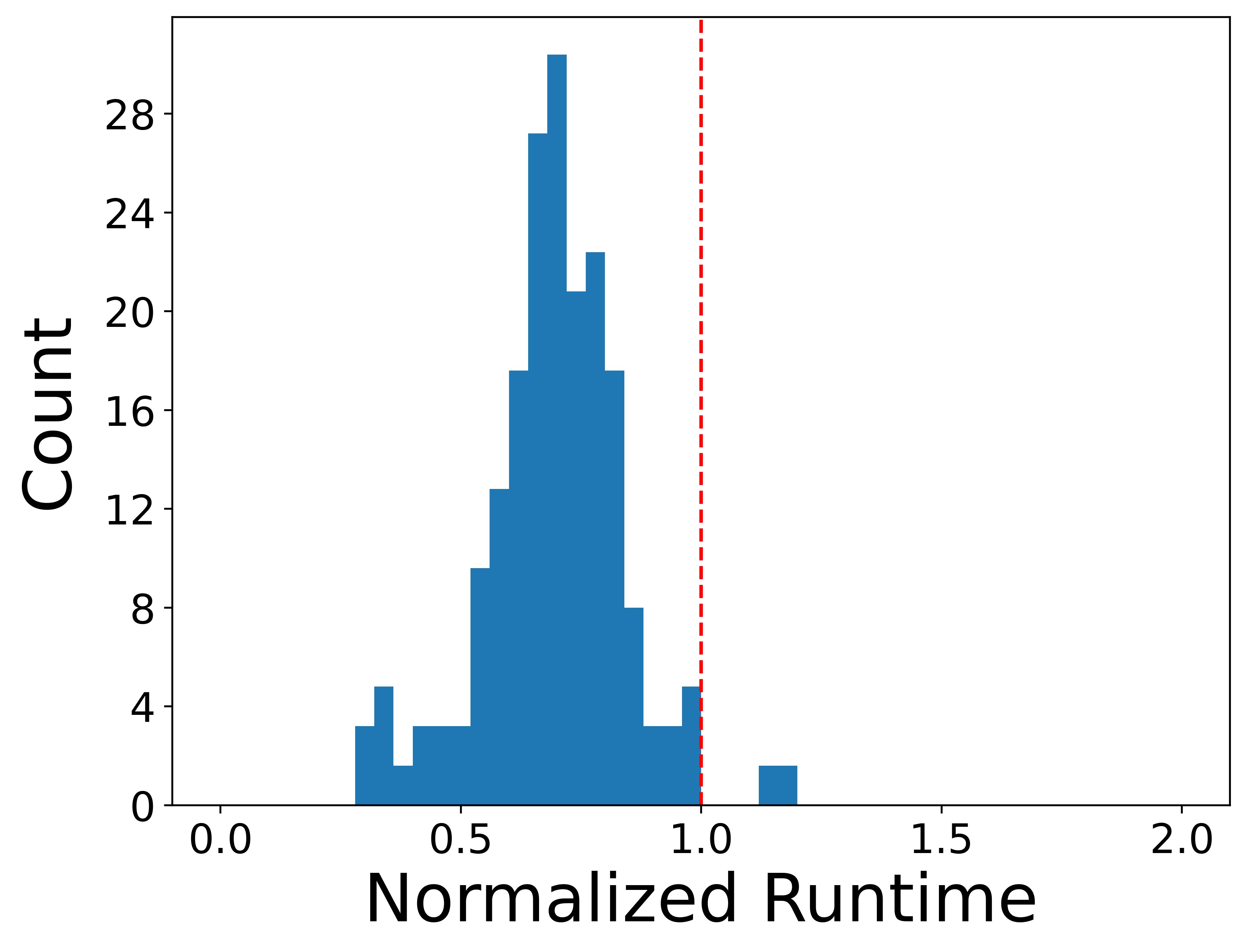}
     \caption{FC}
     \label{fig:d}
 \end{subfigure}
 \hfill
 \begin{subfigure}{0.24\textwidth}
     \includegraphics[width=\textwidth]{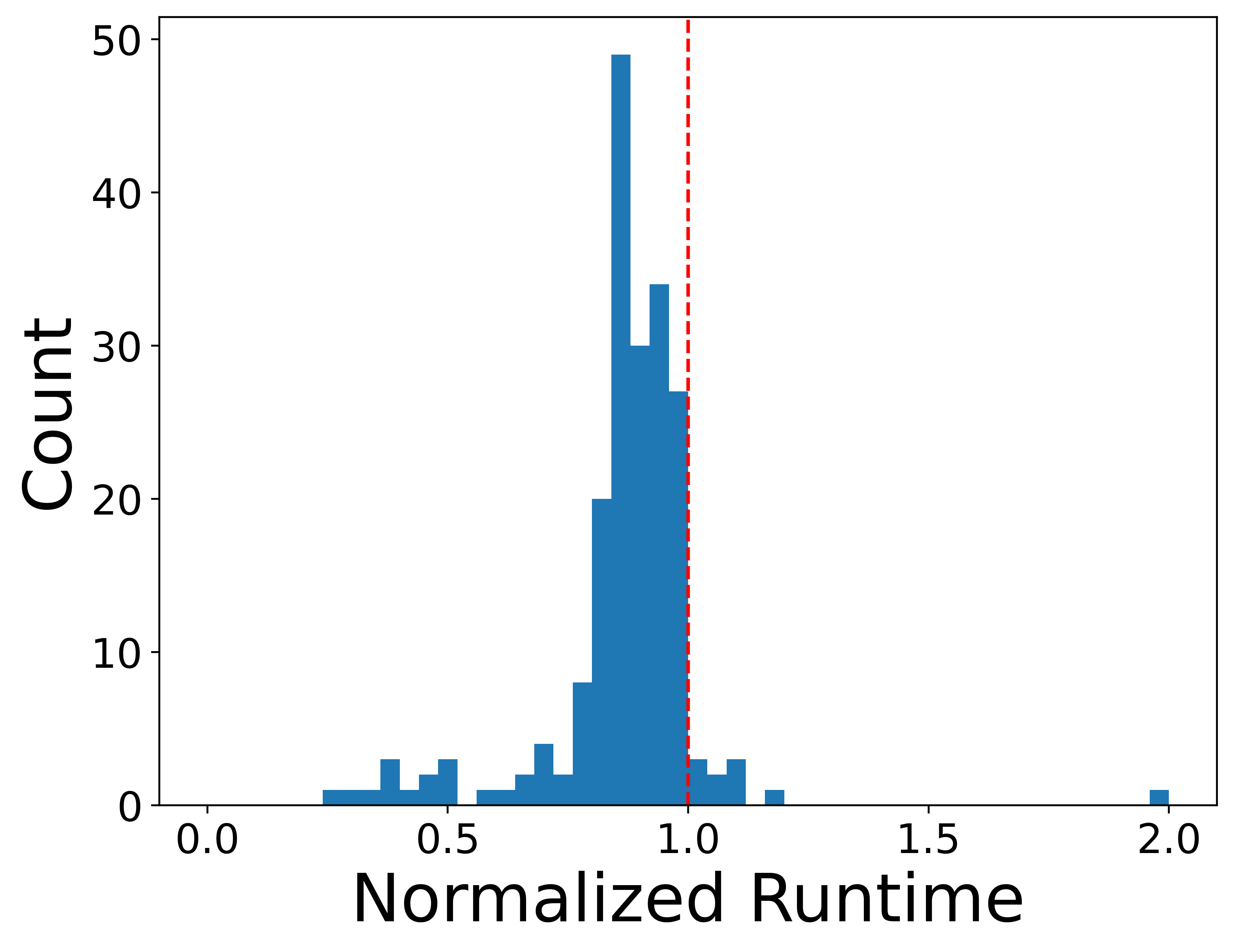}
     \caption{NN}
     \label{fig:c}
 \end{subfigure}
 \\
 \caption{Histograms illustrating the normalized runtime distributions of best candidate backdoors among each instance for six distinct problem domains: (a) GISP-S, (b) SC-S, (c) CA-S, (d) IS-S, (e) FC, and (f) NN. The normalized runtime is calculated as the ratio of the solve time with the candidate backdoor to the original solve time without it. The red vertical line at 1.0 marks the threshold where the candidate backdoor's performance equals the original solve time. Values to the left of this line indicate instances where the candidate backdoor resulted in a faster solve time, while values to the right indicate a slower solve time than the original.}
  \label{fig:bd}
\end{figure}

\begin{figure*} [t]
 \begin{subfigure}{0.49\textwidth}
     \includegraphics[width=\textwidth]{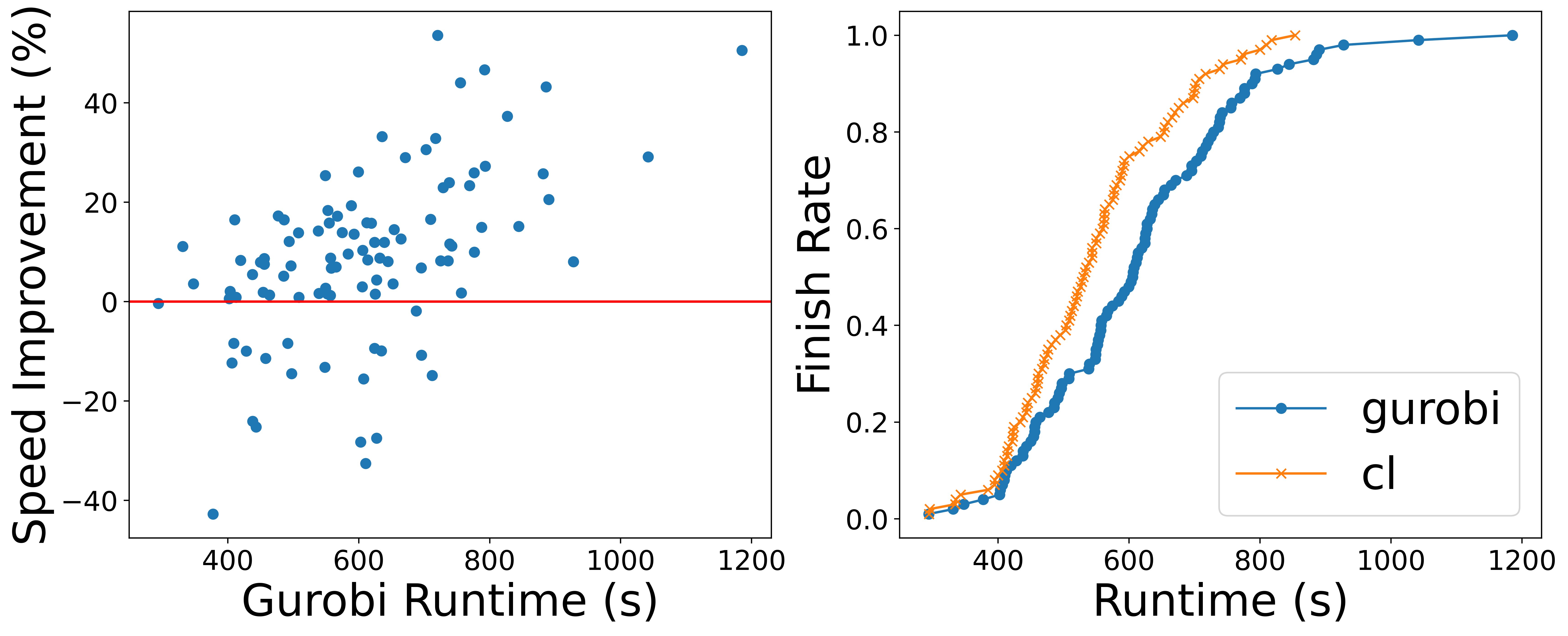}
     \caption{GISP-S}
     \label{fig:a}
 \end{subfigure}
 \hfill
 \begin{subfigure}{0.49\textwidth}
     \includegraphics[width=\textwidth]{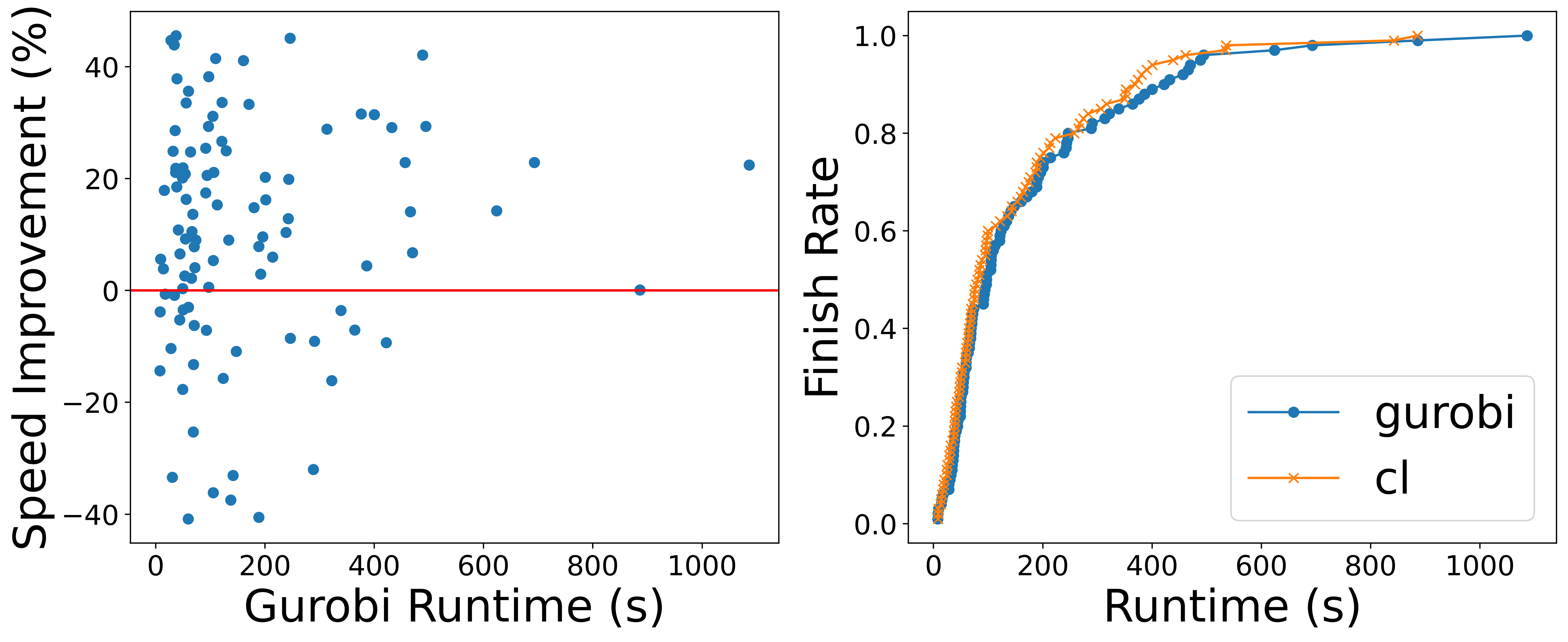}
     \caption{SC-S}
     \label{fig:b}
 \end{subfigure}
 \\
 \\
 \begin{subfigure}{0.49\textwidth}
     \includegraphics[width=\textwidth]{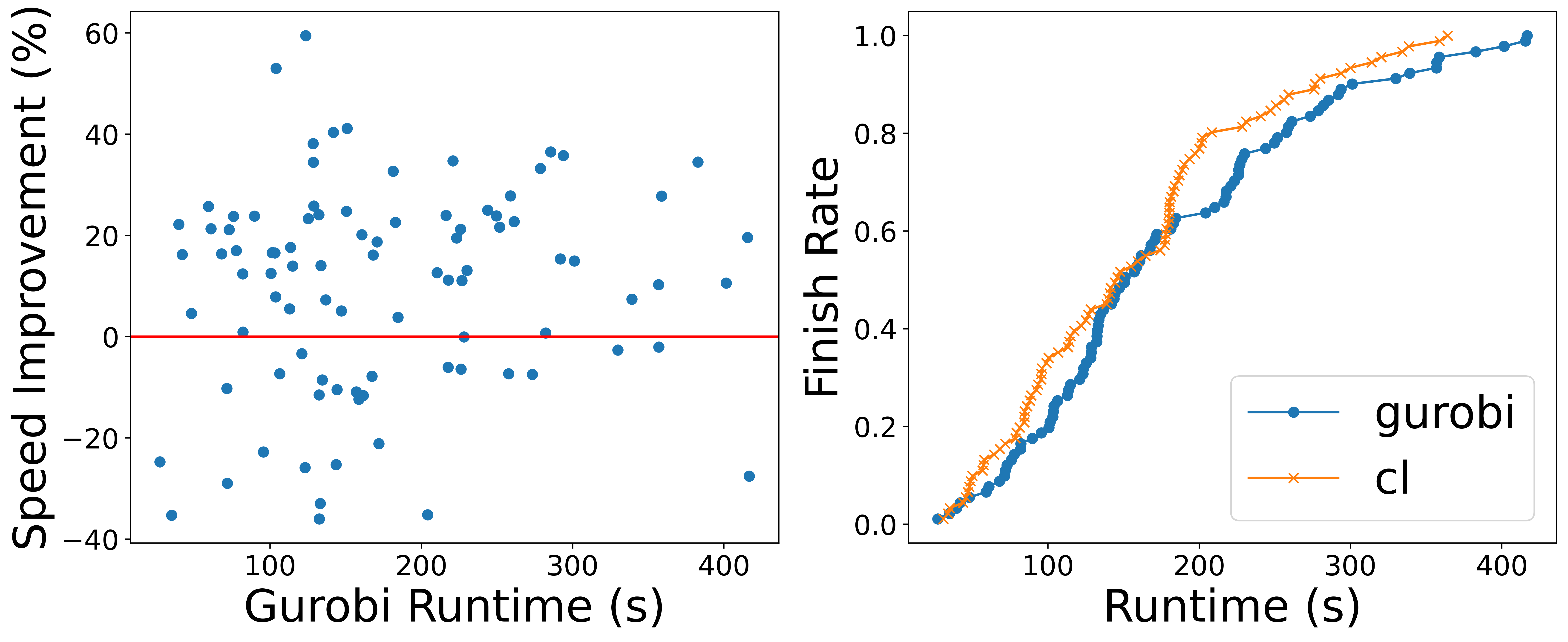}
     \caption{CA-S}
     \label{fig:c}
 \end{subfigure}
 \hfill
 \begin{subfigure}{0.49\textwidth}
     \includegraphics[width=\textwidth]{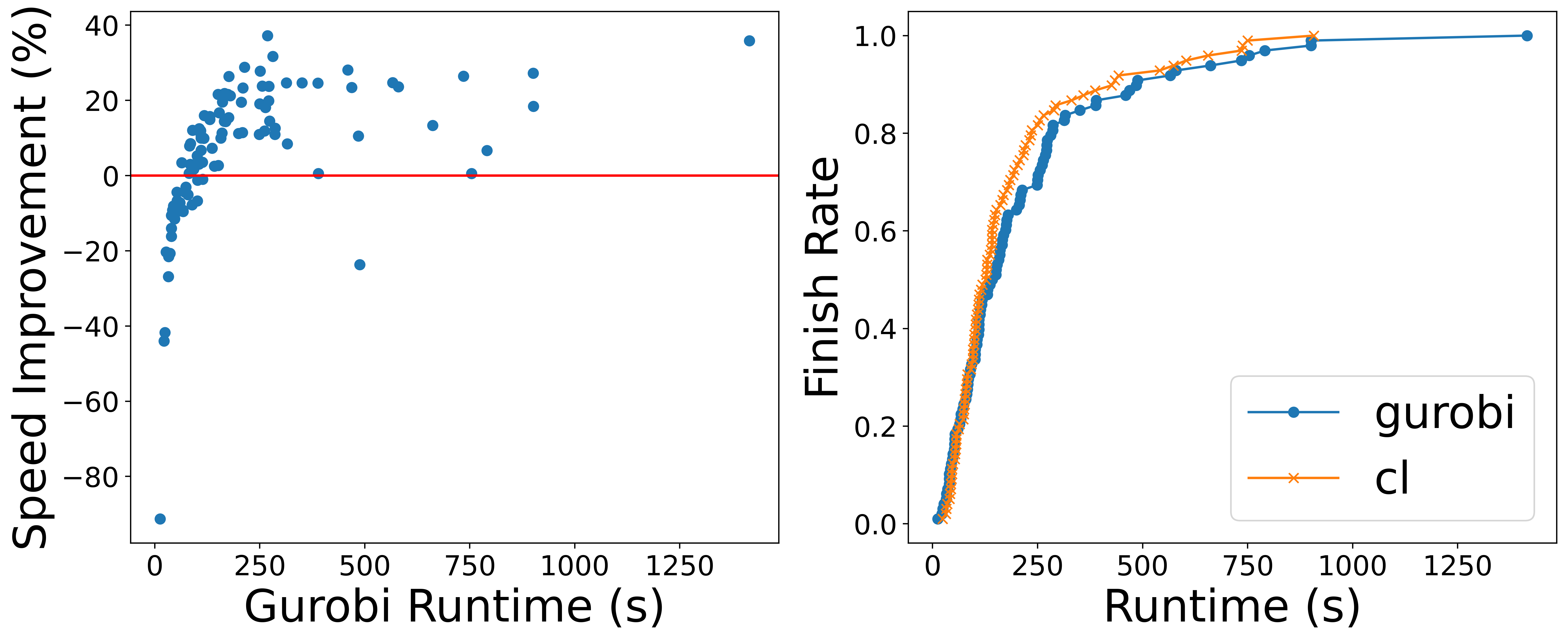}
     \caption{MIS-S}
     \label{fig:d}
 \end{subfigure}
 \\
 \caption{This figure shows the runtime of Gurobi (\textsc{grb}) and contrastive learning model (\textsc{cl}) on GISP-S, SC-S, CA-S, and MIS-S through two different types of plots. The left part is a scatter plot with Gurobi runtime as the x-axis and the speed improvement as the percentage of \textsc{cl} over \textsc{grb} as the y-axis. The points above the red line are ones where \textsc{cl} is better than \textsc{grb} and vice versa. The right part shows the finish rate as a function of runtime. The finish rate for a given runtime is the fraction of instances solved to optimality within the runtime. The blue line is \textsc{grb} and the orange line is \textsc{cl}, with every dot indicating one finished instance. The figures show that \textsc{cl} outperforms \textsc{grb} on average and specifically provides speedups on the harder instances in each distribution.}
  \label{fig:dd}
\end{figure*}

\end{document}